\documentclass{article}
\usepackage{graphicx} 
\usepackage{hyperref}

\usepackage{eurosym}
\usepackage{amsfonts}
\usepackage{amsmath}
\usepackage[left=2cm, right=2cm, top=2cm, bottom=2cm]{geometry}
\usepackage{amssymb}
\usepackage{mathrsfs} 
\usepackage{physics}
\usepackage{siunitx}
\usepackage[dvipsnames]{xcolor}
\usepackage{tikz}

\usepackage{amsthm}
\theoremstyle{plain}

\newtheorem{prop}{Proposition}
\usepackage{caption}
\usepackage{physics}
\newcommand{\fr}{\text{Fr}}
\newcommand{\ro}{\text{Ro}}

\newcommand{\id}{\,\mathrm{d}}
\newcommand{\sdiff}{\,\circ\!\mathrm{d}}

\usepackage{titling}
\usepackage{graphicx}
\usepackage{duckuments}
\usepackage{caption}
\usepackage[onehalfspacing]{setspace}
\usepackage{hyperref}
    \hypersetup{
        colorlinks=true,
        citecolor=Green,
        linkcolor=black,
        filecolor=magenta,      
        urlcolor=black,
        }
\usepackage{algorithm}
\usepackage{algorithmic}
\DeclareMathOperator*{\argmin}{argmin}
\usepackage{subcaption}
\usepackage{booktabs}

\usepackage[mathlines]{lineno}
\usepackage[color=yellow, colorinlistoftodos]{todonotes}
\usepackage{framed}
\usepackage[titletoc, title]{appendix}
    
\usepackage{scalerel}
\usepackage{pict2e}
\usepackage{tkz-euclide}
\usepackage{anyfontsize}
\usetikzlibrary{calc}
\usetikzlibrary{patterns, arrows.meta, decorations.pathreplacing}
\usetikzlibrary{shadows}
\usetikzlibrary{external}

\usepackage{pgfplots}
\pgfplotsset{compat=newest}
\usepgfplotslibrary{statistics}
\usepgfplotslibrary{fillbetween}

\definecolor{WaterBlue}{rgb}{0.11, 0.01 , 0.88}

\graphicspath{ {./figures/} }

\title{Generative Modelling of Stochastic Rotating Shallow Water Noise }
\author{Alexander Lobbe, Dan Crisan, Oana Lang \\
\\ {\small Department of Mathematics, Imperial College London, SW7 2AZ, UK}}
\date{}

\begin{document}

\maketitle

\abstract{In recent work \cite{us}, the authors have developed a generic methodology for calibrating the noise in fluid dynamics stochastic partial differential equations where the stochasticity was introduced to parametrize subgrid-scale processes. The stochastic parameterization of sub-grid scale processes is required in the estimation of uncertainty in weather and climate predictions, to represent systematic model errors arising from subgrid-scale fluctuations. The methodology in \cite{us} used a principal component analysis (PCA) technique based on the ansatz that the increments of the stochastic parametrization are normally distributed.
	In this paper, the PCA technique is replaced by a generative model technique. This enables us to avoid imposing additional constraints on the increments. The methodology is tested on a stochastic rotating shallow water model with the elevation variable of the model used as input data. The numerical simulations show that the noise is indeed non-Gaussian. The generative modelling technology gives good RMSE, CRPS score and forecast rank histogram results. 
}

\section{Introduction}

 Stochastic parameterizations address the uncertainty stemming from unaccounted for or neglected physical effects, as well as inaccurate   observational data and imperfect theoretical models. Over the past two decades, there has been significant research in the area of stochastic parameterizations, largely driven by their application in quantifying uncertainty generated by downsampling high-resolution solutions to lower resolutions. More recently, numerous stochastic parameterization approaches have emerged to tackle such challenges, see for instance \cite{Wei2}, \cite{EtienneLong}, \cite{Holm2015}, \cite{Wei1}, \cite{Memin2014}.

The accurate calibration of the stochastic model parameters can be used in the application of stochastic models, for example, in data assimilation and forecasting processes. Recently, several numerical techniques for calibration (\cite{Wei1}, \cite{EtienneLong}, \cite{Wei2}, \cite{Resseguir2021}) have been developed to demonstrate the effective integration of data-driven models and advanced data assimilation methods. In such studies, the calibration algorithms  typically involve computing the full trajectories of the corresponding fluid parcels, which is often expensive numerically. The approach we introduced in \cite{us} operates with entire solution fields instead. The methodology accounts for small-scale effects which are unresolved as a result of working with models run at coarse resolution, and it uses a principal component analysis (PCA) technique that relies on
the ansatz that the data is Gaussian. However, the Gaussian assumption may not be exactly fulfilled in practice. 

In this paper, we replace the PCA technique with a generative model one, a technical change which allows us to model closer to the data by relaxing the Gaussian assumption. As with
the previous work, we aim to design data-driven models in which real
uncertainty is accounted for, based on input from measurements and
statistically-informed initial data.

We give next a brief description of the stochastic parametrization framework and the calibration methodology introduced in \cite{us}. 
We denote by $m^{f}$ the {deterministic} model state and assume that the
evolution of $m^{f}$ is governed by a partial differential equation of the following form
\begin{equation}
\frac{dm^{f}}{dt}=\mathcal{A}(m^{f}),\ \ t\geq 0,  \label{det1}
\end{equation}%
where $\mathcal{A}$ is the model operator. Given that we implement the equation numerically,  let us assume that the partial differential equation (\ref{det1}) is discretised
in time and space and that the evolution of $m^{f}$ satisfies 
\begin{equation}
m_{t_{n+1}}^{f}=m_{t_{n}}^{f}+\mathcal{A}\left( m_{t_{n}}^{f}\right) \Delta ,
\label{det2}
\end{equation}%
where $0\leq t_{1}\leq t_{2}\leq ...$ is an equidistant time grid with mesh $%
\Delta $. Higher order numerical schemes are possible: the same procedure
will apply to those.  We will denote by $m^{c}$ the (discretised)
stochastic model thought of as being modelled on a coarser spatial grid than that
on which  $m^{f}$ is simulated, hence the superscript $c$. The aim of the
stochastic parametrization is to  compensate for the loss of scales when
moving from a fine grid to a coarse grid. The effect of the \textit{%
unresolved scales} can be mathematically modelled by a term of the form 
\begin{equation}
\sum_{i=1}^{M}\mathcal{M}(m_{t_{n}}^{c})\xi_{i}\sqrt{\Delta }W_{t_{n}}^{i},
\label{det}
\end{equation}
where $\mathcal{M}$ is a suitably chosen operator and $M$ is the number of
sources of noise, $\left( \xi _{k}\right) _{k=1}^{M}$ are (space dependent
but time independent) vector fields and $W_{t_{n}}$ are independent normally
distributed random variables  $W_{t_{n}}^{k}\sim N\left( 0,1\right) $.\footnote{It is this assumption that will be removed in the current study.} 
In
other words, we have    
\begin{equation}
m_{t_{n+1}}^{c}=m_{t_{n}}^{c}+\mathcal{A}\left( m_{t_{n}}^{c}\right) \Delta
+\sum_{k=1}^{M}\mathcal{M}(m_{t_{n}}^{c})\xi _{k}W_{t_{n}}^{k}\sqrt{\Delta }
\label{det3}
\end{equation}
The choice of the stochastic parametrization (\ref{det}) is such that
asymptotically, as $\Delta $ tends to 0, one deduces that the model run on
the coarse grid approximates the stochastic partial differential equation (SPDE)
\begin{equation}
dm^{c}=\mathcal{A}(m^{c})dt+\mathcal{M}(m^{c})dW_{t},\ \ t\geq 0,
\label{sto}
\end{equation}%
where $W_{t}=W(t,x)$ is a space-time Brownian motion. We note that,
theoretically, the solutions of both the deterministic equation (\ref{det1})
and its stochastic counter-part (\ref{sto}) live on the same physical domain
(such as $\mathbb{R}^{n}$, the torus, a horizontal strip, etc), however their time 
\emph{discretisations } \eqref{det2} and \eqref{det3} are approximated on
different space grids and the space discretisation for \eqref{det2} is finer
than the one for \eqref{det3}. Therefore, in the numerical resolution for \eqref{det2} and \eqref{det3}, then we could distinguish between the model
operator for \eqref{det2}, and call it, say $\mathcal{A}^{f}$ and that for 
\eqref{det3}, and call it, say, $\mathcal{A}^{c}$. 

In \cite{us} we estimated the number of sources of noise $M$ and the space
dependent vector fields $\left( \xi _{k}\right) _{k=1}^{N}$ by using a principal component analysis methodology. Obviously, this hinged on the
assumption that the stochastic parametrization that models the small scale
dynamics has Gaussian increments. In practice this may not always be the case. In
this paper we lift this assumption and assume that effect of the \textit{%
unresolved scales} is mathematically modelled by a term of the form 
\begin{equation}
\mathcal{M}(m_{t_{n}}^{c})N_{n}  \label{det4}
\end{equation}%
where $\left( N_{n}\right) _{n\geq 1}$ are independent indentically
distributed random variables, but not necessarily with a Gaussian
distribution. Again, just as in \cite{us}, we estimate the distribution of the independent noises from data and the calibration
procedure introduced below is agnostic to the source of the input data. The \textit{data} can be real data, such as satellite observations of e.g. ocean
sea-surface height, data from re-analysis such as ERA5 (\cite{longlist}), or
synthetic data from a model run of (\ref{det}) computed on a sufficiently
large time window $[0,T]$.  The use of a coarser grid computation in
subsequent data assimilation of model reduction will lead to a significant reduction of computational effort.

Generative models are a class of
machine learning models designed to generate new data samples from an
unknown distribution. They are trained on a given dataset of samples from
the same distribution. An important class of generative models are \textit{%
diffusion models} which have gained more attention recently due to their
ability to generate high-quality and diverse samples. The core idea behind 
\textit{diffusion models }is to iteratively transform the training data
through a diffusion-like mechanism into samples from a known distribution (a
Gaussian distribution for example). In the process, the forward and the
backward diffusions are learned using a neural network. Once the learning is complete, samples from the unknown distribution are obtained by running the backward diffusion initiated from samples from the Gaussian distribution. We give details of the methodology we use which is based on
\cite{de2021diffusion} in Section 3, specifically tailored to calibrate a stochastic rotating
shallow water model. 

In this paper we use synthetic data coming from a realization of the
(deterministic) rotating shallow water model, for which we keep the same
notation $m^{f}$ for now. The model run is then mollified using a procedure
that will eliminate the small/fast scales effects, for example by using a
low-pass filter, Gaussian mollifier, Helmholtz projection, subsampling, etc, or combinations thereof. We will denote by $C(m^{f})$ the resulting
mollification of the data. Note that both $m^{f}$ and $C(m^{f})$ live on the
same space. We emphasise that $C(m^{f})$ is not the solution of (\ref{sto})
and its time discretisation will not satisfy (\ref{det3}). However, we make
the ansatz that the difference between the two processes $\hat{m}%
:=m^{f}-C(m^{f})$ has a stochastic representation given by (\ref{det4}), in
other words, we will have 
\begin{equation}
\hat{m}_{t_{n+1}}-\hat{m}_{t_{n}}\approx \displaystyle\mathcal{M}%
(m_{t_{n}}^{c})N_{n},  \label{diff2}
\end{equation}{where }$N_{n}${\ has an unknown distribution to be modeled by
a certain }score-based generative model known as a diffusion Schr\"{o}dinger bridge,
{following~\cite{de2021diffusion}}. Details of our implementation are included in
the Section \ref{sec:diffusion-models}.

{In the context of stochastic modeling in fluid dynamics, uncertainty plays
a significant role, and accurately calibrating these models to real-world
scenarios is crucial for reliable predictions. In this paper we show that
diffusion models can be used to quantify the uncertainty due to \textit{%
unresolved scales}}. The end result is that we can {generate ensembles of
fluid states. These fluid models can be affected by uncertainty coming from
other sources not just from unresolved scale, see \cite{berner} for details} . For example, one
may want to model fast scales through a stochastic parametrization. This is in
line with the Hasselmann paradigm (see \cite{chk}) where a
stochastic model of climate variability entails slow changes of climate that are explained as the integral response to continuous random excitation by short period "weather" disturbances. Therefore the model will incorporate a
rapidly varying "weather" system (essentially the atmosphere) modelled
\textit{stochastically}, and a slowly responding "climate" system (the ocean,
cryosphere, land vegetation, etc.) modelled \textit{deterministically}. The essential feature of stochastic climate models is that the non-averaged "weather" components are also retained. They appear formally as stochastic forcing terms. {\ Calibrating stochasticity that models fast scales is different. In
this case, the "truth" is already stochastic (the stochasticity is part of
the model) and the data is made out of increments of the truth -
minus the drift term. The low pass filter is not used here as the stochasticity is not a result of the coarsening procedure. However, we can still apply the generative model approach to infer the stochastic terms. In this
case, the original model is in fact stochastic:} 
\begin{equation}
dm=\mathcal{A}(m)dt+\mathcal{M}(m)dW_{t},\ \ t\geq 0.  \label{sto2}
\end{equation}%
As a result 
\begin{equation}
m_{t_{n+1}}-m_{t_{n}}-\mathcal{A}\left( m_{t_{n}}\right) \Delta \approx %
\displaystyle\mathcal{M}(m_{t_{n}})N_{n}.  \label{det5}
\end{equation}%
In other words, the data consists of the increments $m_{t_{n+1}}-m_{t_{n}}-%
\mathcal{A}\left( m_{t_{n}}\right) \Delta $ out of which we compute the
samples from the distribution of $N_{n}$.

In the following subsection we provide an overview of the contents of the paper.

\subsection{Outline of the Paper}

In Section~\ref{sec:shallow-water} we describe the particular fluid dynamics model we will be working with throughout the paper. Specifically this is a rotating shallow water (RSW) model, similar to the model used in the earlier works (\cite{us}, {\color{red}...}). The novelty here compared to previous works is that we use a non-dimensionalised version of the rotating shallow water model, whose derivation is briefly outlined in Section~\ref{sec:shallow-water}. We choose to use a non-dimnensionalised model in order to be able to change the physical properties of the flow based on the selection of the non-dimensional numbers: the Rossby and Froude numbers in this case. Given that we are modelling subgrid scale effects by noise, it is instrumental that the coarse and fine scale deterministic model evolutions are sufficiently different, to ensure that there is a clear target for the noise term. 

In Section~\ref{sec:diffusion-models} we introduce the generative model used in the numerical studies of this paper. The theoretical framework allows for different types of generative models, some of which we mention in the beginning of Section~\ref{sec:diffusion-models} below. We choose to use the  \emph{diffusion Schr\"odinger bridge} model because it is a promising candidate for the fluid modelling studies we perform for several reasons. Specifically, the model is relatively transparent from a mathematical point of view, due to the form of the mathematically derived diffusion model. This is very expressive as a machine learning model, because of the underlying parametric model which is a neural network. Finally, we think that the \textit{Schr\"odinger bridge} is useful due to the iterative nature that should make the calibration of the number of diffusion steps less critical.

In Section~\ref{sec:numerics} we present the numerical study and results we have obtained based on the non-dimensionalised rotating shallow water equations. We verify that the evolution of the fluid we simulate indeed exhibits a loss of scales. Next, we show that the non-standard dataset we use to train the diffusion Schr\"odinger bridge is indeed representable by the parametric model. Further, the stochastic ensemble run with the generative model is shown to have an advantage compared to Gaussian noise in terms of different forecast metrics. The RMSE and CRPS scores significantly improve when the generative noise is used in the low initial uncertainty setting.  

In Section~\ref{sec:conclusion} we summarise the conclusions of our study and identify directions for future studies.

\section{Rotating Shallow Water Model}
\label{sec:shallow-water}
In this paper, we base our study on a stochastic approximation of the nondimensionalised rotating shallow water model

\begin{equation}\label{rsw}
    \begin{aligned}
     \id_t\vb{u} + (\vb{u}\cdot\grad) \vb{u} + \frac{f}{\ro} \hat{\vb{z}}\times\vb{u}  
     +\frac{1}{\fr^2} \grad (\eta - b)&=0\\
     \id_t\eta + \div{(\eta\vb{u})}  &=0
    \end{aligned}
\end{equation}
where
\begin{itemize}
    \item $\vb{u}(x,t) = (u(x,t), v(x,t))$ is the horizontal fluid velocity vector field
    \item $\eta(x,t)$ is the height of the fluid column
    \item $f \in \mathbb{R}$ is the Coriolis parameter, $f=2\Theta \sin\varphi$ where $\Theta$
    is the rotation rate of the Earth and $\varphi$ is the latitude; $f\hat{z}
    \times \vb{u} = (-fv, fu)^T$, where $\hat{z}$ is a unit vector pointing away
    from the centre of the Earth
    \item $\fr = \frac{U}{\sqrt{gH}}$ is the Froude number (dimensionless) which is connected to the stratification of the fluid flow. Here $U$ is a typical scale for horizontal speed and $H$ is the typical vertical scale, while $g$ is the gravitational acceleration.
    \item $\ro = \frac{U}{f_0 L}$ is the Rossby number (also dimensionless) which describes the effects of rotation on the fluid flow: a small Rossby number ($\ro << 1$) suggests that the rotation term dominates over the advective terms.
    \item $b(x,t)$ is the bottom topography function.
     \end{itemize}
The initial condition is computed from an initial $\eta$-field from a  geostrophic balance assumption (see details in Section \ref{subsect:finevscoarse}). We work with the corresponding discrete version of (\ref{rsw}), that is
\begin{equation}\label{rswd11}
    \begin{aligned}
     \vb{u}_{n+1}-\vb{u}_{n} + (\vb{u}_n\cdot\grad) \vb{u}_n\Delta + \frac{f}{\ro} \hat{\vb{z}}\times\vb{u}_n\Delta  
     +\frac{1}{\fr^2} \grad (\eta_n - b)\Delta&=0\\
     \eta_{n+1}-\eta_{n} + \div{(\eta_n\vb{u}_n)}\Delta  &=0
    \end{aligned}
\end{equation}

In \cite{us} we perturbed the iteration corresponding to (\ref{rswd11}) with spatial Gaussian noise of the form 
\begin{equation}\label{noise}
W_n(x)=\sqrt{\Delta} \sum_{i=1}^{M} \boldsymbol{\xi}_i(x) W_n^i
    \end{equation}
where $(\boldsymbol{\xi}_i)_i$ are divergence-free elements of the covariance basis functions of the SALT (\cite{Holm2015}) noise parametrisation and  $W^i_n\sim N(0,1)$ are independent i.i.d. random variables. When we do this, we obtain the following recurrence formula
\begin{equation}\label{rswd}
    \begin{aligned}
     \vb{u}_{n+1}-\vb{u}_{n} + (\tilde{\vb{u}}_n\cdot\grad) \vb{u}_n + \vb{u}_n \cdot \grad W_n(x) + \frac{f}{\ro} \hat{\vb{z}}\times\tilde{\vb{u}}_n  
     +\frac{1}{\fr^2} \grad (\eta_n - b)\Delta &=0\\
     \eta_{n+1}-\eta_{n} + \div{(\eta_n\tilde{\vb{u}}_n)}  &=0
    \end{aligned}
\end{equation}
where \footnote{The term $\tilde{\tilde{\vb{u}}}$ is a velocity perturbation which is specific for this stochastic version of the RSW model.}
\begin{equation}\label{pert}
\tilde{\vb{u}}_n=\vb{u}_n \Delta + W_n(x) 
\end{equation}
The choice of the perturbation (\ref{pert}) is such that the iteration (\ref{rswd}) is an approximation of the stochastic partial differential equation 
\begin{equation}
    \begin{aligned}
     \id\vb{u} + \left[(\vb{u}\cdot\grad) \vb{u} + \frac{f}{\ro} \hat{\vb{z}}\times\vb{u}\right] \id t
    + \sum_i \left[(\boldsymbol{\xi}_i \cdot \grad) \vb{u} + \grad\boldsymbol{\xi}_i\cdot\vb{u} + \frac{f}{\ro} \hat{\vb{z}}\times\boldsymbol{\xi}_i\right]\sdiff W_t^i 
    &= -\frac{1}{\fr^2} \grad (\eta - b)\id t\\
     \id\eta + \div{(\eta\vb{u})} \id t + \sum_i \div{(\eta\boldsymbol{\xi}_i)}\sdiff W_t^i &=0
    \end{aligned}
\end{equation}
where $\circ$ denotes Stratonovich integration and $W^i$ are standard i.i.d. Brownian motions as before. The Stratonovich stochastic term generates a second order correction when writing the system in It\^{o} form, but this is dealt with using the intrinsic properties of the numerical scheme. We have explained this part in detail in the Appendix of \cite{us}.

To bring the rotating shallow water example in line with the general notation presented in the introduction, observe that $m^f$ is represented here by the pair $(\vb{u}, \eta)$ which solves the partial differential equation (\ref{rsw}). Since we will be working only with discrete approximations, we can directly identify $m^f$ with the solution of (\ref{rswd11}).  Then $m^c$ is the solution of (\ref{rswd}). In other words:
\[
m_{t_n}^c:=\left( 
\begin{array}{c}
\vb{u}_{t_n}^c \\ 
\eta_{t_n}^c
\end{array}
\right)
\]
where $(\vb{u}_{t_n}^c, \eta_{t_n}^c)$ solves \eqref{rswd}. 
Then
\begin{equation}\label{opm}
\mathcal{M}(m_{t_n}^c)(\zeta) = \mathcal{M}\left( 
\begin{array}{c}
\vb{u}_{t_n}^c \\ 
\eta_{t_n}
\end{array}
\right)(\zeta) :=\left( 
\begin{array}{c}
\grad \vb{u}_{t_n}^c \cdot \zeta + u_{t_n}^c \cdot \grad \zeta\\ 
\grad \eta_{t_n}^c \cdot \zeta.
\end{array}
\right)
\end{equation}
with $\zeta$ typically corresponding to $\xi_i$, \footnote{Here we can observe one more time the challenges posed by transport noise in general (and SALT noise in this particular case) as this always involves calculating derivatives corresponding to both the model variable $m$ and the (noise) variable $\zeta$. In other words, the operator $\mathcal{M}$ and the variable $\zeta$ are inherently intertwined.} 
and
\begin{equation}
\mathcal{A}(m_{t_n}^c) = \mathcal{A}\left( 
\begin{array}{c}
\vb{u}_{t_n}^c \\ 
\eta_{t_n}^c
\end{array}
\right) =\left( 
\begin{array}{c}
(\vb{u}_{t_n}^c\cdot\grad) \vb{u}_{t_n}^c + \frac{f}{\ro} \hat{\vb{z}}\times\vb{u}_{t_n}^c  
     +\frac{1}{\fr^2} \grad (\eta_{t_n}^c - b)  \\ 
\div{(\eta_{t_n}^c\vb{u}_{t_n}^c)}
\end{array}
\right).
\end{equation}
Based on \cite{us} we have
\begin{equation}\label{eq:increments} 
\hat{m}_{t_{n+1}} - \hat{m}_{t_n}
 \approx \displaystyle\sum_{i=1}^M\mathcal{M}(m_{t_n}^c)\xi_i(x)\sqrt{\Delta} W_{t_n}^i = \mathcal{M}(m_{t_n}^c) W_{t_n}(x)
\end{equation}
with $\mathcal{M}$ given above in \eqref{opm} and
\begin{equation}
W_{t_n}(x)=\sqrt{\Delta} \sum_{i=1}^{M} \boldsymbol{\xi}_i(x) W_{t_n}^i
\end{equation}
for the particular case of the RSW model.
In practice, however, we first generate the increments in \eqref{eq:increments} for $\eta$ only and then we use them together with a geostrophic balance assumption to compute the corresponding noise increments for the two components of $\vb{u}_{t_n}^c$. That is, for the RSW model we work mainly with $\mathcal{M}(\eta_{t_n}^c)(\zeta) = (\grad \eta_{t_n}^c \cdot \zeta)$ which corresponds to a \textit{transport noise}. 
The novelty of the current work is that we replace the spatial Gaussian noise $W_{t_n}(x)$ in \eqref{diff2} with a general noise $N_n(x)$ 
such that
\begin{equation}\label{diffnew}
\hat{m}_{t_{n+1}}-\hat{m}_{t_{n}}\approx \displaystyle\mathcal{M}%
(m_{t_{n}}^{c})N_{n}  
\end{equation}
{where }$N_{n}$ has an unknown distribution which is modelled using a diffusion Schr\"{o}dinger bridge.

\begin{figure}[H]
    \centering
    \includegraphics[width=0.8\textwidth]{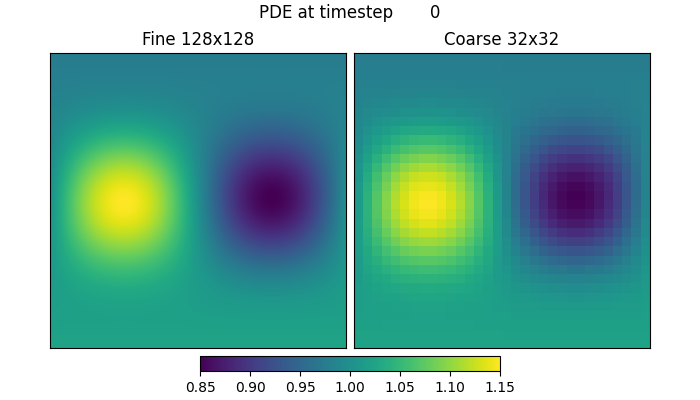}
    \caption{Initial Condition for the non-dimensional height variable on the fine (128x128) and coarse (32x32) grids}
    \label{fig:init-cond}
\end{figure}

\begin{figure}[H]
    \centering
    \includegraphics[width=0.8\textwidth]{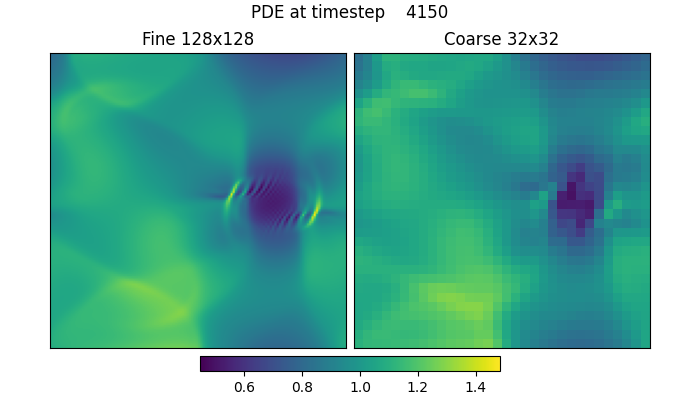}
    \caption{Snapshot of non-dimensional height variable at fine and coarse run after 4150 fine scale timesteps. We can observe the development of fine scale features which are absent (unresolved) in the coarse field.}
    \label{fig:fine-scale-waves}
\end{figure}

To ensure that the noise $N_n$ is divergence-free, we generate a scalar random field $\tilde N_n$ such that 
$N_n=\nabla^\perp \tilde N_n$.  In other words,  $\tilde N_n$ sastisfies a hyperbolic partial differential equation
\begin{equation}
\hat{m}_{t_{n+1}}-\hat{m}_{t_{n}}= - C(m_{t_{n}}^{f})\grad^\perp\tilde N_{n}  
\end{equation}
where $C$ is a low-pass filter as defined in \cite{us} Section 3. 
We refer to Section 3 in \cite{us} for the detailed procedure we use to solve this 
partial differential equation.

\section{Score-Based Generative Models: Diffusion Schr\"odinger Bridge}
\label{sec:diffusion-models}
Score-based generative models are a recent trend in generative modelling and have achieved state-of-the-art results in several benchmark tasks in Machine Learning (cite some). In general,  generative models are used to generate previously unseen samples from an underlying probability distribution which is typically available only through a dataset of samples. The generation of new samples is performed by a learned model which represents the unknown distribution of the training data either implicitly or explicitly. Classical examples of generative models in the literature include, among many others, Gaussian Mixture Models (GMMs), Hidden Markov Models (HMMs), Autoregressive Models~\cite{pmlr-v48-oord16}, Variational Autoencoders (VAEs)~\cite{kingma2013auto}, Generative Adversarial Networks (GANs)~\cite{goodfellow2014generative}, Energy-Based Models (EBMs)~\cite{lecun2006tutorial}, Normalising Flows~\cite{papamakarios2021normalizing} and, most recently, Diffusion Generative Models~\cite{ho2020denoising, sohl2015deep, song2019generative, song2021scorebased, de2021diffusion}. A comprehensive review of the field of generative modelling with diffusion models is provided in~\cite{yang2023diffusion}.
Note that, since the data distribution is typically a distribution over a very high-dimensional state-space, neural networks are the standard choice for the underlying parametric model in nearly all generative models used nowadays. 

Although there are several different types of diffusion models in the literature, they all follow a common principle. That is, the data is being gradually diffused by successively adding noise (we call this below a \textit{noising process}) until it becomes essentially a sample from a pure noise distribution, such as a Gaussian with known mean and covariance. Then, one can reverse the noising process to generate samples from the unknown data distribution by drawing a pure noise sample and running the reverse diffusion process. This is called the \emph{denoising} process.

In this work, we are focusing on a specific type of score-based diffusion known as the Diffusion Schr\"odinger Bridge (DSB) model, developed in~\cite{de2021diffusion}.
Broadly speaking, the DSB model is an extension of the classical score-based diffusion model by an optimal transport procedure known as iterative proportional fitting (IPF) which allows to iterate the score-based diffusion training and can work with shorter noising and denoising processes as a result. In the following subsection we outline the basics of the DSB model following~\cite{de2021diffusion}.

\subsection{Learning Diffusion Schr\"odinger Bridge}

The noising process is modelled by a forward Markov Chain  $\{X_k\}_{k=0}^N$ on $\mathbb{R}^d$ such that
\begin{equation*}
    X_{k+1} = X_k + \gamma_{k+1} f(X_k) + 2\gamma_{k+1} V_{k+1},\;\;\; k=0,\dots, N,
\end{equation*}
where $\{V_{k+1}\}_k \sim \mathcal{N}(0,\mathbf{1})$ are i.i.d.~Gaussian random variables, $f: \mathbb{R}^d\to\mathbb{R}^d$ is a drift function and $\{\gamma_k\}_k$ are typically small stepsize parameters.  The symbol $\mathbf{1}$ denotes the identity matrix. We assume the initial density $X_0\sim p_0 = p_{data}$. The joint density $p$ of the Markov Chain $X_{0:N}=(X_0, \dots, X_N)$ can be decomposed into the corresponding forward transition densities $\{p_{k+1 | k}\}_k$ as follows
\begin{equation*}
    p(x_{0:N}{ {=\{x_{\scaleto{k}{3pt}}\}_{ \scaleto{k=0}{3pt}}^{\scaleto{N}{3pt}}\in(\mathbb{R}^{\scaleto{d}{3pt}})^{\scaleto{N+1}{3pt}}}}) = p_0(x_0)\prod_{k=0}^{N-1} p_{k+1 | k}(x_{k+1}| x_k).
\end{equation*}
Similarly we can write down the backward decomposition
\begin{equation*}
    p(x_{0:N}) = p_N(x_N)\prod_{k=0}^{N-1} p_{k | k+1}(x_{k}| x_{k+1}) 
    = p_N(x_N)\prod_{k=0}^{N-1} \frac{p_k(x_k)p_{k+1|k}{(x_{k+1}|x_k)}}{p_{k+1}(x_{k+1})},
\end{equation*}
where $\{p_k\}_k$ are the marginal densities and $\{p_{k|k+1}\}_k$ are the reverse transition densities.
The methodology is based  on sampling from $p_\text{data}$ using the reverse decomposition initialized at $p_N=p_\text{prior}$. To achieve this, we need to approximate the reverse transition densities. To this end note the earlier assumption that the backward transitions are normally distributed as 
\begin{equation*}
    p_{k+1|k}(x_{k+1}|x_k) = \mathcal{N}(x_{k+1}; x_k+\gamma_{k+1}f(x_k), 2\gamma_{k+1}\mathbf{1})
\end{equation*}
 Then we apply a Taylor approximation (see~\cite{de2021diffusion}) to get
\begin{equation*}
\begin{aligned}
p_{k \mid k+1}\left(x_k \mid x_{k+1}\right) 
 &=p_{k+1 \mid k}\left(x_{k+1} \mid x_k\right) \exp \left[\log p_k\left(x_k\right)-\log p_{k+1}\left(x_{k+1}\right)\right] \\
& \approx \mathcal{N}\left(x_k ; x_{k+1}-\gamma_{k+1} f\left(x_{k+1}\right)+2 \gamma_{k+1} \nabla \log p_{k+1}\left(x_{k+1}\right), 2 \gamma_{k+1} \mathbf{1}\right)
\end{aligned}
\end{equation*}
Then, the backward transitions are Gaussian, with a drift depending on the parameters $f$ and $\{\gamma_k\}_k$ and on the \emph{score functions} $\{\nabla \log p_{k}\}_k$. Note that we can integrate out the initial density from the marginals such that
\begin{equation*}
    p_{k+1}(x_{k+1}) = \int p_0(x_0)p_{k+1|0}(x_{k+1}|x_0) dx_0
\end{equation*}
and thus
\begin{equation*}
    \nabla \log p_{k+1}(x_{k+1}) = \mathbb{E}_{p_{0|k+1}}[\nabla_{x_{k+1}} \log p_{k+1|0}(x_{k+1}|X_0)].
\end{equation*}
The conditional expectation above is intractable, but the joint distribution is available through samples, so we can use regression to find it
    \begin{equation*}
s_{k+1}=\arg \min _s \mathbb{E}_{p_{0, k+1}}\left[\left\|s\left(X_{k+1}\right)-\nabla_{x_{k+1}} \log p_{k+1 \mid 0}\left(X_{k+1} \mid X_0\right)\right\|^2\right],
\end{equation*}
where $\|\cdot\|$ denotes the $L_2$-norm on $\mathbb{R}^d$.
We can thus learn a parametrised approximation of the score (all scores simultaneously)
\begin{equation*}
    s_{\theta^\star}(k,x_k) \approx \nabla \log p_k(x_k)
\end{equation*}
via Denoising Score Matching (Vincent 2011) as
\begin{equation*}
    \theta^\star = \arg\min_\theta \sum_{k=1}^N \mathbb{E}_{p_{0,k}}[\|s_\theta(k,X_k) -\nabla_{x_k} \log p_{k|0}(X_k|X_0) \|^2].
\end{equation*}
So we estimate the score function and then sample $X_0\overset{\scriptscriptstyle approx}{\sim} p_{data}$ using the diffusion started at $p_N\approx p_{prior}$ such that
\begin{equation*}
    X_k = X_{k+1} - \gamma_{k+1}f(X_{k+1}) + 2\gamma_{k+1}s_{\theta^\star}(k+1, X_{k+1}) + \sqrt{2\gamma_{k+1}} \mathcal{N}(0,\mathbf{1}).
\end{equation*}
Let $\mathcal{P}_{N+1}$ be the space of sequences of probability densities of length $N+1$. In the Schr\"odinger Bridge framework, we consider the joint density $p\in\mathcal{P}_{N+1}$ of the Markov Chain $X$ and we want to find a density $\pi^\star \in \mathcal{P}_{N+1}$ such that
\begin{equation}\label{eq:IPF-objective}
\pi^{\star}=\argmin \left\{\mathrm{KL}(\pi \mid p): \pi \in \mathcal{P}_{N+1}, \pi_0=p_{\text {data }}, \pi_N=p_{\text {prior }}\right\},
\end{equation}
where for any two probability densities $p$ and $q$ over a space $\mathcal{X}$, 
$$KL(p || q) = \int_\mathcal{X} p(x) \log\left(\frac{p(x)}{q(x)}\right) dx
$$ denotes the Kullback-Leibler divergence\footnote{Note that the Kullback-Leibler divergence is not a distance in a strict mathematical sense because it is not symmetric. Hence the name divergence.} between probability distributions.
Assuming $\pi^{\star}$ is available, a generative model can be obtained by sampling $X_N \sim p_{\text {prior }}$, followed by the reverse-time dynamics $X_k \sim \pi_{k \mid k+1}^{\star}\left(\cdot \mid X_{k+1}\right)$ for $k \in\{N-1, \ldots, 0\}$.

A well-known solution method to find a minimum of \ref{eq:IPF-objective} is iterative proportional fitting (IPF). Initialised at $\pi^0 = p(x_{0:N})$ this method defines the following iterative process
\begin{equation*}
\begin{aligned}
& \pi^{2 n+1}=\argmin \left\{\mathrm{KL}\left(\pi \mid \pi^{2 n}\right): \pi \in \mathcal{P}_{N+1}, \pi_N=p_{\text {prior }}\right\} \\
& \pi^{2 n+2}=\argmin \left\{\mathrm{KL}\left(\pi \mid \pi^{2 n+1}\right): \pi \in \mathcal{P}_{N+1}, \pi_0=p_{\text {data }}\right\}
\end{aligned}
\end{equation*}
A positive result for the feasibility of IPF in our setting is provided in Proposition~\ref{prop:IPF} below.
\begin{prop}[Proposition 2 in~\cite{de2021diffusion}]\label{prop:IPF}
Assume that $\mathrm{KL}\left(p_{\text {data }} \otimes p_{\text {prior }} \mid p_{0, N}\right)<+\infty$. Then for any $n \in \mathbb{N}, \pi^{2 n}$ and $\pi^{2 n+1}$ admit positive densities w.r.t. the Lebesgue measure denoted as $p^n$ resp. $q^n$ and for any $x_{0: N} \in \mathcal{X}$, we have $p^0\left(x_{0: N}\right)=p\left(x_{0: N}\right)$ and
$$
q^n\left(x_{0: N}\right)=p_{\text {prior }}\left(x_N\right) \prod_{k=0}^{N-1} p_{k \mid k+1}^n\left(x_k \mid x_{k+1}\right), p^{n+1}\left(x_{0: N}\right)=p_{\text {data }}\left(x_0\right) \prod_{k=0}^{N-1} q_{k+1 \mid k}^n\left(x_{k+1} \mid x_k\right)
$$
\end{prop}

In practice we have access to $p_{k+1 \mid k}^n$ and $q_{k \mid k+1}^n$. Hence, to compute $p_{k \mid k+1}^n$ and $q_{k+1 \mid k}^n$ we use
$$
p_{k \mid k+1}^n\left(x_k \mid x_{k+1}\right)=\frac{p_{k+1 \mid k}^n\left(x_{k+1} \mid x_k\right) p_k^n\left(x_k\right)}{p_{k+1}^n\left(x_{k+1}\right)}, q_{k+1 \mid k}^n\left(x_{k+1} \mid x_k\right)=\frac{q_{k \mid k+1}^n\left(x_k \mid x_{k+1}\right) q_{k+1}^n\left(x_{k+1}\right)}{q_k^n\left(x_k\right)}.
$$
The following Proposition~\ref{prop:loss} details a possible loss function to use in the training of the DSB model, known as Mean Matching. Different variations to be used for training can be found in~\cite{de2021diffusion}.
\begin{prop}[Proposition 3 in~\cite{de2021diffusion}]
\label{prop:loss}
Assume that for any $n \in \mathbb{N}$ and $k \in\{0, \ldots, N-1\}$,
$$
q_{k \mid k+1}^n\left(x_k \mid x_{k+1}\right)=\mathcal{N}\left(x_k ; B_{k+1}^n\left(x_{k+1}\right), 2 \gamma_{k+1} \mathbf{I}\right), p_{k+1 \mid k}^n\left(x_{k+1} \mid x_k\right)=\mathcal{N}\left(x_{k+1} ; F_k^n\left(x_k\right), 2 \gamma_{k+1} \mathbf{I}\right),
$$
with $B_{k+1}^n(x)=x+\gamma_{k+1} b_{k+1}^n(x), F_k^n(x)=x+\gamma_{k+1} f_k^n(x)$ for any $x \in \mathbb{R}^d$. Then we have for any $n \in \mathbb{N}$ and $k \in\{0, \ldots, N-1\}$
$$
\begin{aligned}
& B_{k+1}^n=\argmin _{\mathrm{B} \in \mathrm{L}^2\left(\mathbb{R}^d, \mathbb{R}^d\right)} \mathbb{E}_{p_{k, k+1}^n}\left[\left\|\mathrm{~B}\left(X_{k+1}\right)-\left(X_{k+1}+F_k^n\left(X_k\right)-F_k^n\left(X_{k+1}\right)\right)\right\|^2\right], \\
& F_k^{n+1}=\argmin _{\mathrm{F} \in \mathrm{L}^2\left(\mathbb{R}^d, \mathbb{R}^d\right)} \mathbb{E}_{q_{k, k+1}^n}\left[\left\|\mathrm{~F}\left(X_k\right)-\left(X_k+B_{k+1}^n\left(X_{k+1}\right)-B_{k+1}^n\left(X_k\right)\right)\right\|^2\right] .
\end{aligned}
$$
\end{prop}
Note that, here, we use
 neural networks $ B_{\beta^n}(k, x) \approx B_k^n(x) \text { and } F_{\alpha^n}(k, x) \approx F_k^n(x)$ to parametrise the unknown drifts in the transition densities. 
We describe the DSB training process according to the paper~\cite{de2021diffusion}. For pseudocode of this algorithm, the reader may refer to~\cite{de2021diffusion}. Roughly speaking, the transition densities (drifts) for both the forward and backwards noising and denoising processes are modelled by a collection of neural networks.
The initial joint density $p(x_{0:N})$ is given from a random initialisation of the networks. Each DSB iteration consists of a forward and backward run.
The first step is a forward iteration, where we take samples from the dataset and diffuse them according to the dynamics given by the noising process with the parameterised forward net. 
The losses collected during the forward iterations are applied to the \emph{backward} nets.
In the inner backward iterations we sample from the prespecified prior density (Gaussian) and run the denoising process according to the backward net.
In the backward iterations we apply the gradient descent steps for the \emph{forward} nets.
 The inner iterations are each run until convergence. In practise, a prespecified number of iterations that we tune.

\section{Numerical Results}
In the following we describe our numerical results using the DSB method onn the non-dimensionalised rotating shallow water equations. On a general note, we found that the method of generating noise from a generative model is generally stable and we obtain suitable new noise samples from the trained model. In our setup we generate the noise as an integral of the velocity perturbations used in the SPDE model, akin to a streamfunction in classical GFD. Therefore, the output from the generative model is subject to the application of a gradient. Thus, a small percentage of the data produces large gradients which can lead to instabilities in the numerical evolution of the SPDE. We mitigate this problem by clipping large gradients. Note that this is not expected to be problematic for two reasons. Firstly, the amount of clipped data locations is small and secondly, we are interested in the overall spatial correlations in the generated noise, which remains intact after clipping the gradients because large gradients primarily occur in the boundary regions. 
\label{sec:numerics}
\subsection{Fine vs Coarse Scale}\label{subsect:finevscoarse}

We use an initial height field for $\eta$ given by
\begin{equation*}
    \eta_0(x,y) = 1-\frac{a}{2}\atan(\frac{y}{L_y}-\frac{1}{2}) + a \sin(\frac{2\pi x}{L_x}) + \frac{a}{2}
    \sin(\frac{2\pi x}{L_x})\sin(\frac{\pi y}{L_y} )^4 ,
\end{equation*}
where the domain is the rectangle $[0,L_x]\times [0, L_y]$. We use the domain $[0,1]\times [0, 1]$ in our simulations. Moreover, $a\in\mathbb{R}$ is a parameter. We chose $a=0.1$.
The initial condition for the nondimensional simulation is computed from the initial $\eta$-field using the geostrophic balance condition. This condition in dimensional form for the dimensional velocity component $v^d$ in $y$-direction reads as
\begin{equation*}
    f^dv^d = g\frac{\partial \eta^d}{\partial x^d},
\end{equation*}
where $g=9.81 \,\frac{m}{s^2}$ is the gravitational constant and $f^d$ is a parameter associated with the Coriolis force. The superscript $d$ indicates dimensional variables throughout.
    In non-dimensional form, we have
\begin{equation*}
    f_0 f U v = g \frac{\Delta \eta}{L \Delta x} \;\leftrightarrow\; v = \frac{1}{f}\frac{\text{Ro}}{\text{Fr}^2}\frac{\partial \eta}{\partial x}        
\end{equation*}
where $U$ is a typical velocity scaling, $f_0$ is a typical Coriolis scaling and $L$ is a typical length scale. Here, $\text{Ro}$ denotes the Rossby number and $\text{Fr}$ denotes the Froude number. The simulations are run at $\text{Ro}=0.2$ and $\text{Fr}=1.1$.
Thus also for the non-dimensional velocity component in $x$-direction we have the geostrophic balance condition
\begin{equation*}
    u = -\frac{1}{f}\frac{\text{Ro}}{\text{Fr}^2}\frac{\partial \eta}{\partial y}.
\end{equation*} 
The finally used initial condition is scaled so that the initial $u$ and $v$ variables are $O(1)$ on the domain.

We use the intermediate fields during the spin-up of the system as a qualitative sanity check to verify that the solutions on the fine and coarse grids diverge as the flow progresses. Specifically, we expect to observe fine-scale features (waves) to develop in the fine-grid simulation, which are not resolved in the coarse scale run. Indeed, the plot in Figure~\ref{fig:fine-scale-waves} shows that this is indeed occurring.

\subsection{Training Data}
\begin{figure}[h!]
    \centering
    \begin{subfigure}{0.45\textwidth}\includegraphics[width=\linewidth]{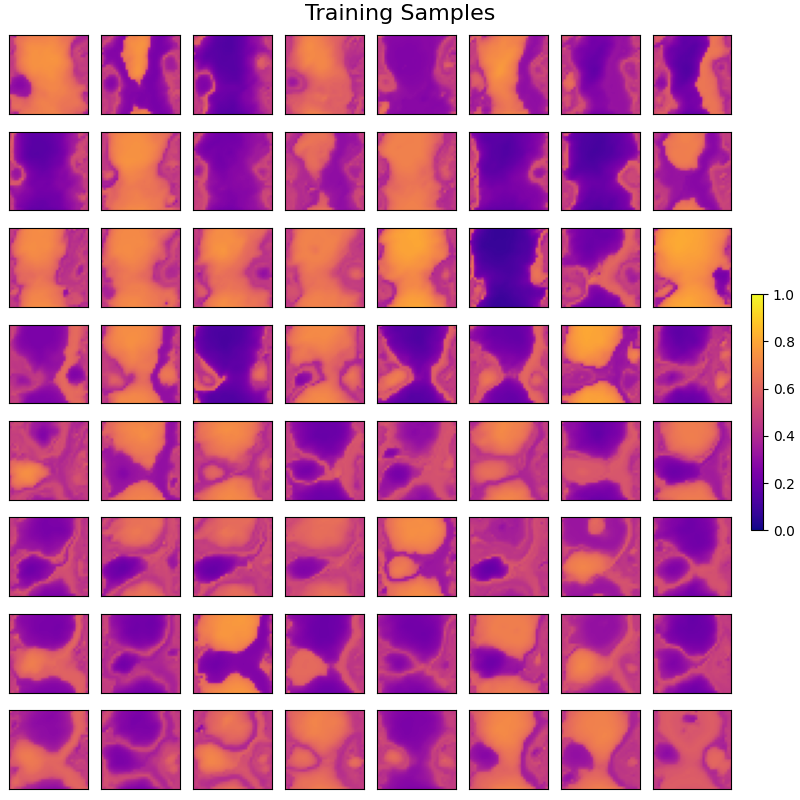}
    \caption{Training Samples}
    \label{fig:training-dataset}
    \end{subfigure}
    \begin{subfigure}{0.45\textwidth}
    \includegraphics[width=\linewidth]{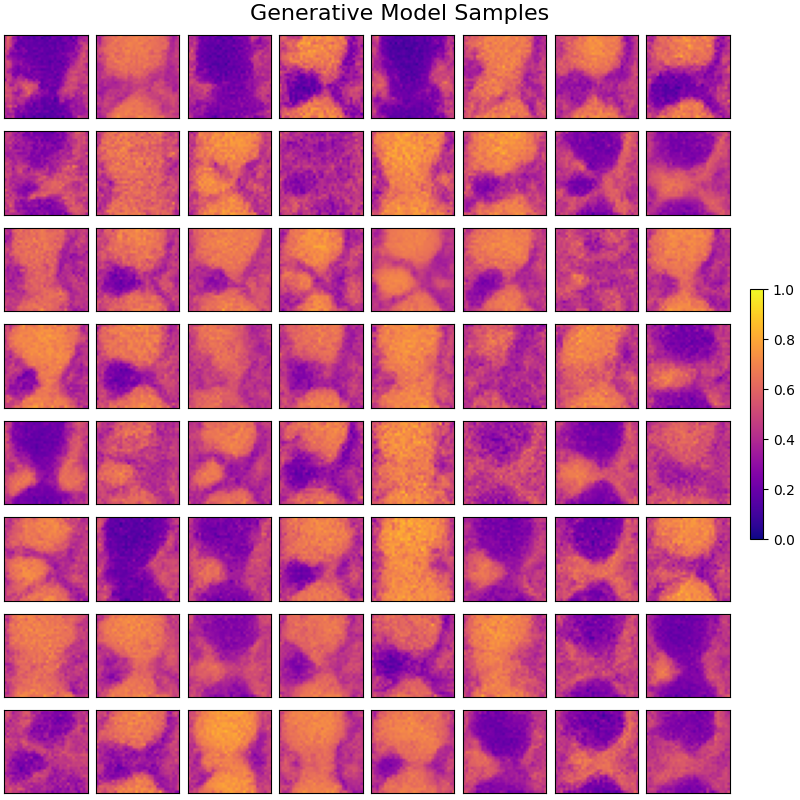}
    \caption{Generated Samples}
    \label{fig:generated-samples}
    \end{subfigure}
    \caption{Training Samples and Samples from the generative model. Samples of the training data after transformation. The fields are outputs of the calibration equation thought of a stream functions for the velocity perturbations in the SPDE. The data have been transformed by a $arcsinh$ transformation and normalized to the interval $[0,1]$. Samples from the generative model.}
\end{figure}

\begin{figure}[h!]
    \centering
    \begin{subfigure}{0.46\textwidth}
    \includegraphics[width=\linewidth]
    {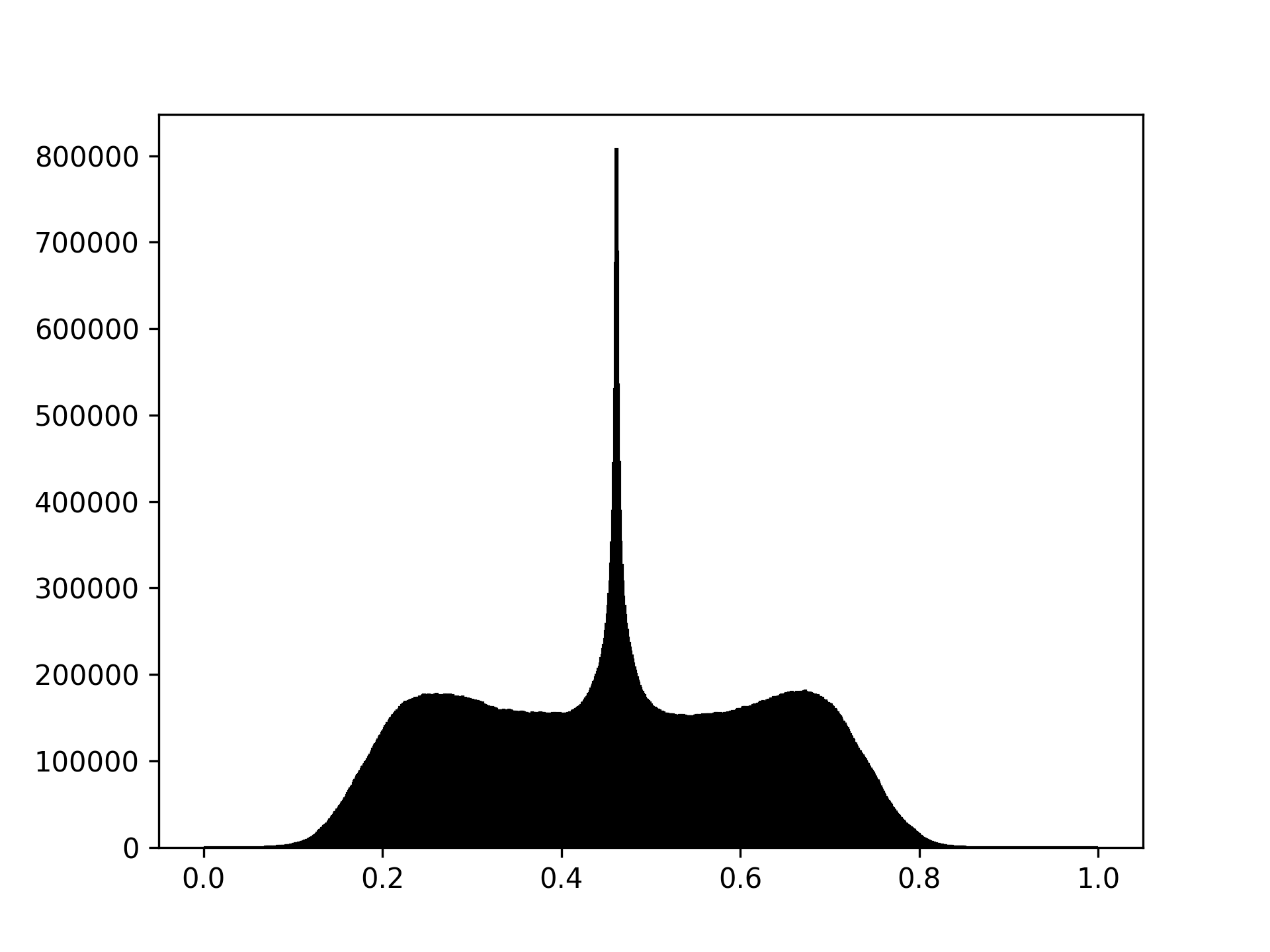}
    \caption{Distribution of pixel values}
    \label{fig:pixel-value-distribution}
    \end{subfigure}   
    \begin{subfigure}{0.40\textwidth}
    \includegraphics[width=\linewidth]{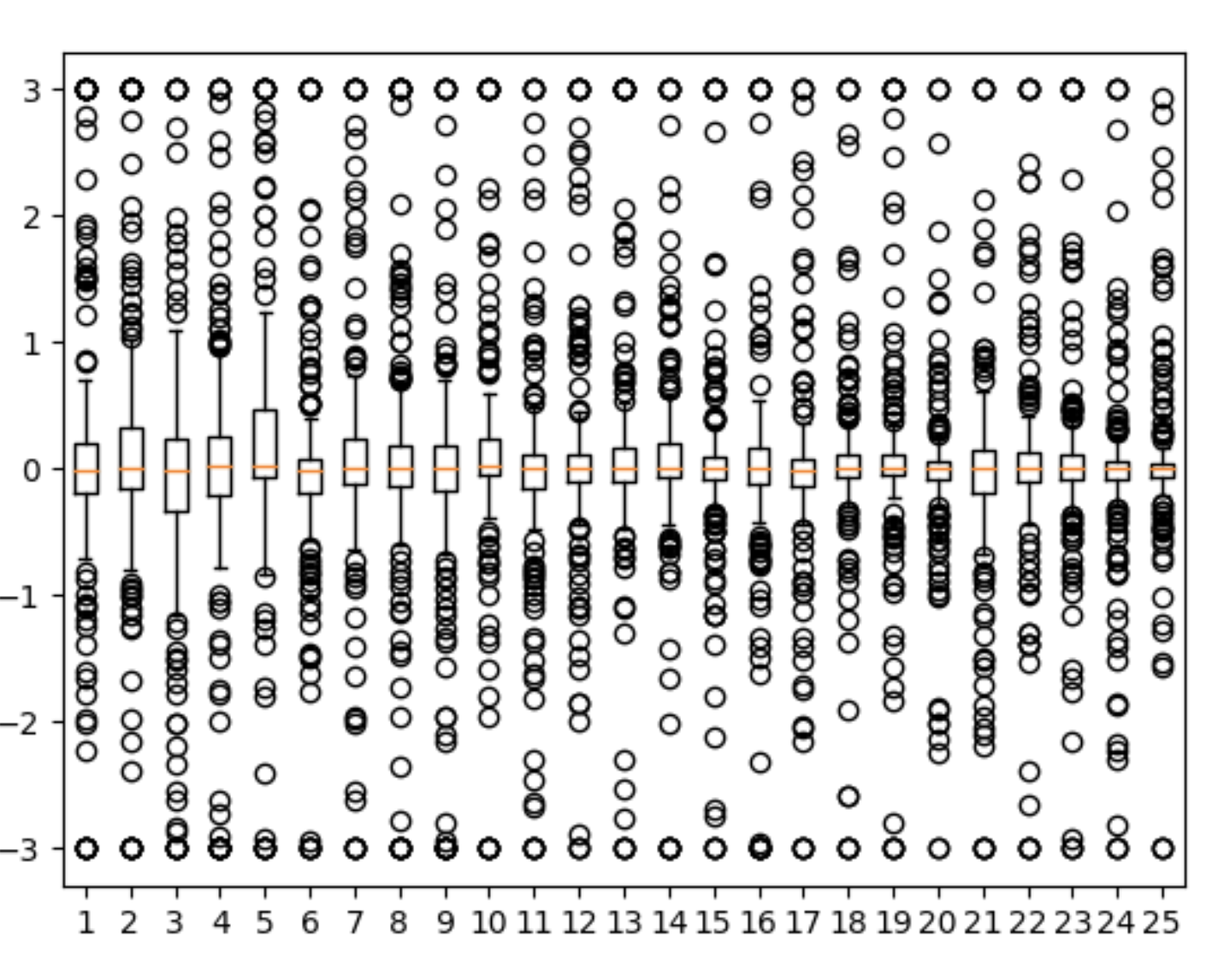}
    \caption{boxplots}
    \label{fig:generated-data-distribution}
    \end{subfigure}
    
    \caption{(a) Distribution of pixelvalues throughout the whole training data set. The data transformation has been choosen s.th. the pixel values achieve a good coverage of the data range $[0,1]$. This way the model can better distinguish between variations in shade.(b) Distribution of pixelvalues in central grid locations in the generated dataset, i.e. the output from the generative model.}
\end{figure}

We generate the training data as solutions of the hyperbolic calibration equation through a forward run of the fine-scale PDE. The collected solutions are thought of as stream functions for the velocity perturbations of the SPDE and are assumed to be sampled from a fixed (in time) probability distribution that we aim to model through the generative model. To ease the training, we perform a nonlinear transformation ($arcsinh$-transform)~\cite{arcsinh}
of the obtained data and normalise the values globally to the interval $[0,1]$. This is standard practise in Machine Learning and Statistics. Specifically, let $\Psi: \Omega\rightarrow\mathbb{R}$ denote a solution of the calibration equation. Then we transform it using
\begin{equation*}
    \hat{\Psi} = \frac{1}{\vartheta} \operatorname{arcsinh}(\vartheta \Psi)
\end{equation*} 
with parameter $\vartheta=2\mathrm{e}5$. The transformed dataset $\{\hat{\Psi}_i\}_i$ is then normalised to the range $[0,1]$ by
\begin{equation*}
    \psi_i = \frac{
    \hat{\Psi}_i -\min_{i,x,y}{\hat{\Psi}_i(x,y)}
    }
    {\max_{i,x,y} (\hat{\Psi}_i(x,y) -\min_{i,x,y}{\hat{\Psi}_i(x,y))}}.
\end{equation*}
Samples from the training set are displayed in Figure~\ref{fig:training-dataset} and the pixel distribution across all samples after transformation and normalisation is depicted in the histogram in Figure~\ref{fig:pixel-value-distribution}.

\subsection{Generative Model Output}

Samples from the generative model output after training are shown in Figure~\ref{fig:generated-samples}. The model has been trained for 5 DSB Iterations using 30 diffusion steps. These samples are subsequently reverse-transformed and the gradients are applied to convert the stream function information into velocity perturbations. The distribution of velocity perturbations (here for the velocity component $u$) at some central grid locations is depicted in the boxplots in Figure~\ref{fig:generated-data-distribution}. It shows that the median of the generated noise distributions at each of the grid locations is close to zero and their magnitude rarely exceeds $1$, as we expect in a non-dimensional simulation. The boxplots also reflect the fact that we clip the noise data at a magnitude of $3$ to avoid instabilities due to outliers. The plot shows that the amount of data clipped is negligible as it is well within the outlier regime at all gridpoints. Moreover, the generative noise is not Gaussian. We perform a one-sample Kolmogorov-Smirnov Test to show that the generated noise is not Gaussian. Specifically, we use the velocity perturbations computed from the generative model noise to inspect the distribution in different spatial locations on the computational grid ($8\times 8=64$ total locations). For each location, we perform independent KS-tests for normality, which show that at none of the tested locations could we detect a normal distribution of the generated noise (p-value less than 0.05).

\subsection{Forecast Studies}
The aim of the generative model is to produce a distribution for the noise in the rotating shallow water model that is advantageous for the modelling of the effects of unrepresented small scales. To this end, we use the established ensemble and forecast metrics root mean square error (RMSE) and continuous ranked probability score (CRPS). We also use rank histograms.
The forecasts are produced from a forward ensemble run of the SPDE with different noises for comparison. The first noise is the generative model noise, which we compare to two different typed of Gaussian noises. One is a Gaussian with the same overall mean and variance as the generative model, i.e. a Gaussian with covariance $\sigma\mathbf{1}$, where $\sigma\in\mathbb{R}$ is the standard deviation of the dataset obtained as the output from the generative model. The second Gaussian noise has a diagonal covariance that is varies in space given as $\operatorname{diag}(\boldsymbol{\sigma})$, where $\boldsymbol{\sigma}\in\mathrm{R}^{mn\times mn}$ is the vector of the standard deviations of the generated noise at all individual spatial locations. The forecasts are run for a lead time of $200$ calibration time steps and then reset to the fine PDE value from which the ensemble is relaunched launched. Additionally, we apply a normally distributed perturbation to each ensemble, to represent initial uncertainty. We chose three different scenarios here, one is a scenario of no initial incertainty, then small initial uncertainty with variance $0.001^2$ and a large initial uncertainty scenario with variance $0.05^2$. 

\begin{figure}
    \centering
    \begin{subfigure}{0.3\textwidth}
    \includegraphics[width=\textwidth]{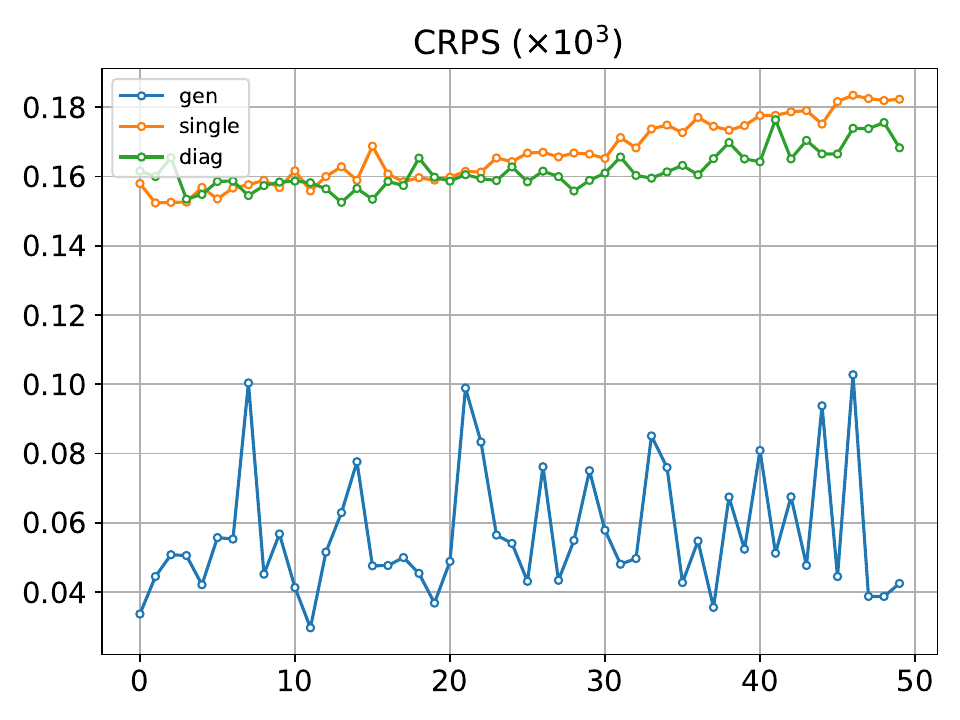}
        \caption{Variable $u$, Std $0.0$}
    \label{fig:crps_200reps_0}
    \end{subfigure}
    \begin{subfigure}{0.3\textwidth}
    \includegraphics[width=\textwidth]{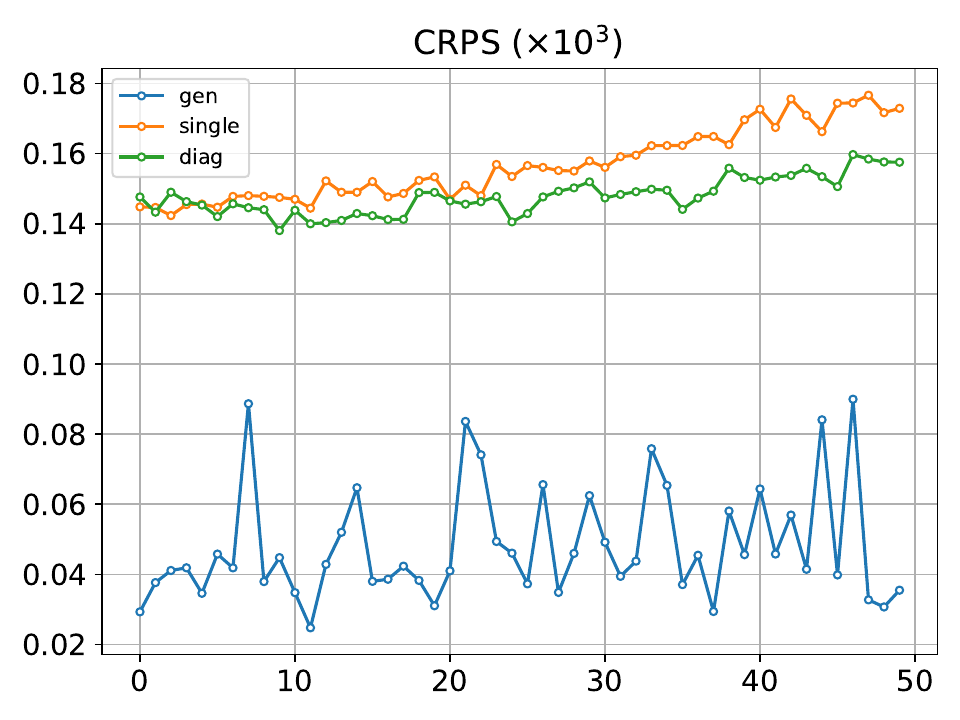}
        \caption{Variable $v$, Std $0.0$}
    \label{fig:crps_200reps_1}
    \end{subfigure}
    \begin{subfigure}{0.3\textwidth}
    \includegraphics[width=\textwidth]{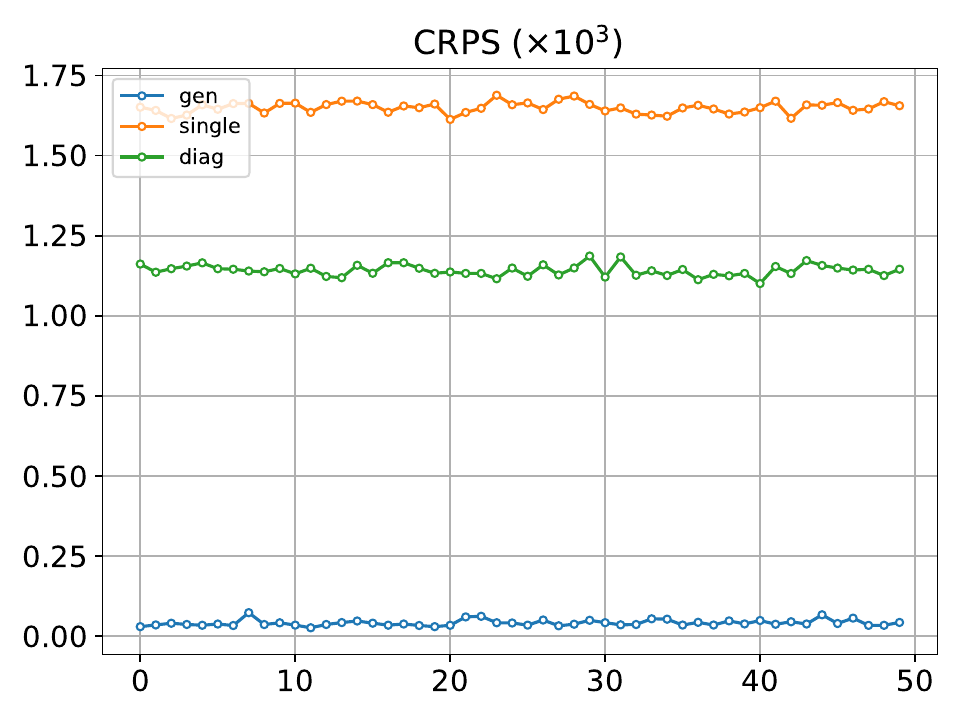}
        \caption{Variable $e$, Std $0.0$}
    \label{fig:rps_200reps_2}
    \end{subfigure}
    \hfill
    \begin{subfigure}{0.3\textwidth}
    \includegraphics[width=\textwidth]{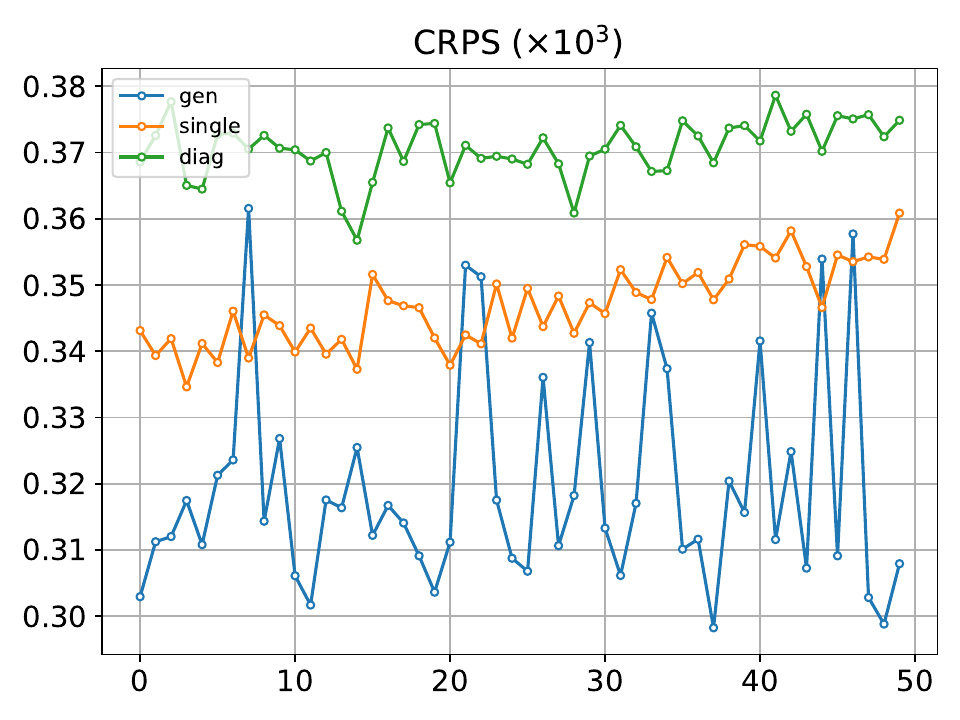}
        \caption{Variable $u$, Std $0.001$}
    \label{fig:crps_200reps_0.001pert_0}
    \end{subfigure}
    \begin{subfigure}{0.3\textwidth}
    \includegraphics[width=\textwidth]{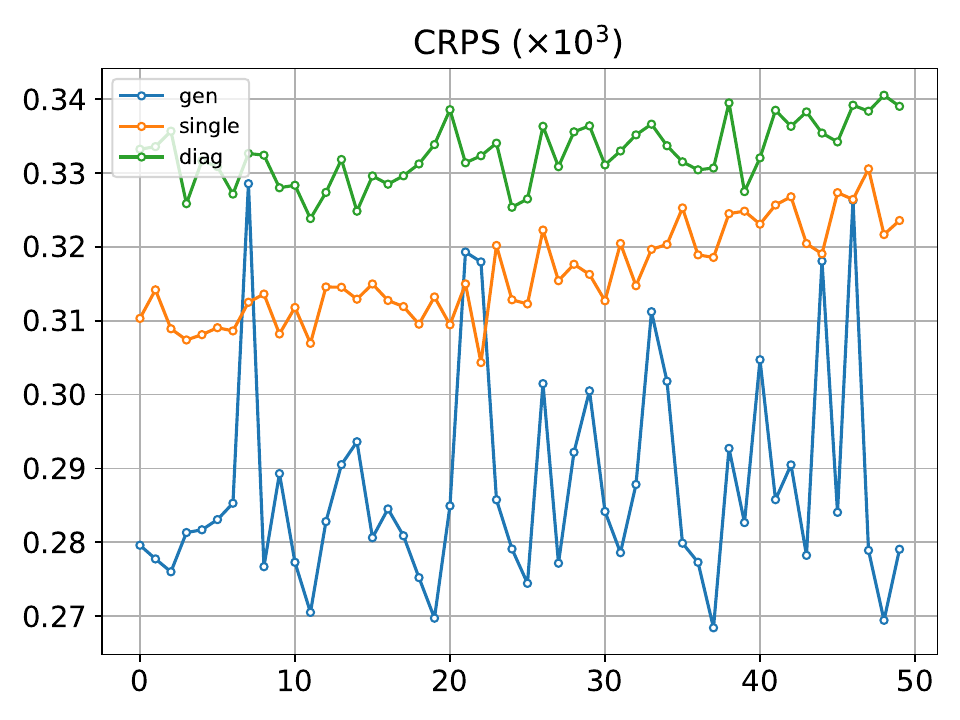}
        \caption{Variable $v$, Std $0.001$}
    \label{fig:crps_200reps_0.001pert_1}
    \end{subfigure}
    \begin{subfigure}{0.3\textwidth}
    \includegraphics[width=\textwidth]{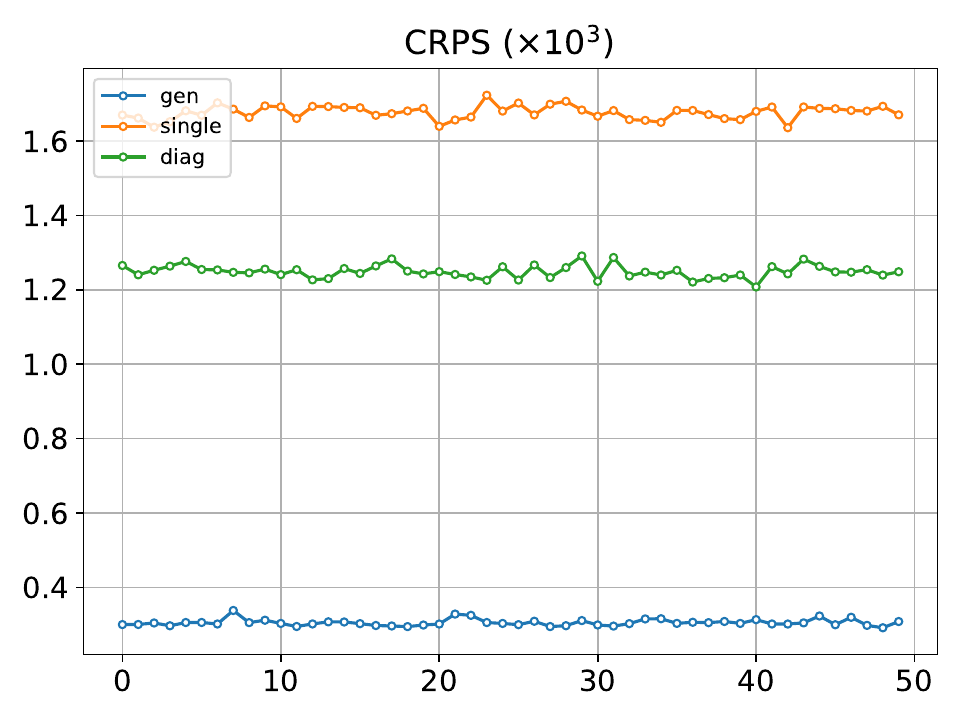}
        \caption{Variable $e$, Std $0.001$}
    \label{fig:crps_200reps_0.001pert_2}
    \end{subfigure}\hfill
    \begin{subfigure}{0.3\textwidth}
    \includegraphics[width=\textwidth]{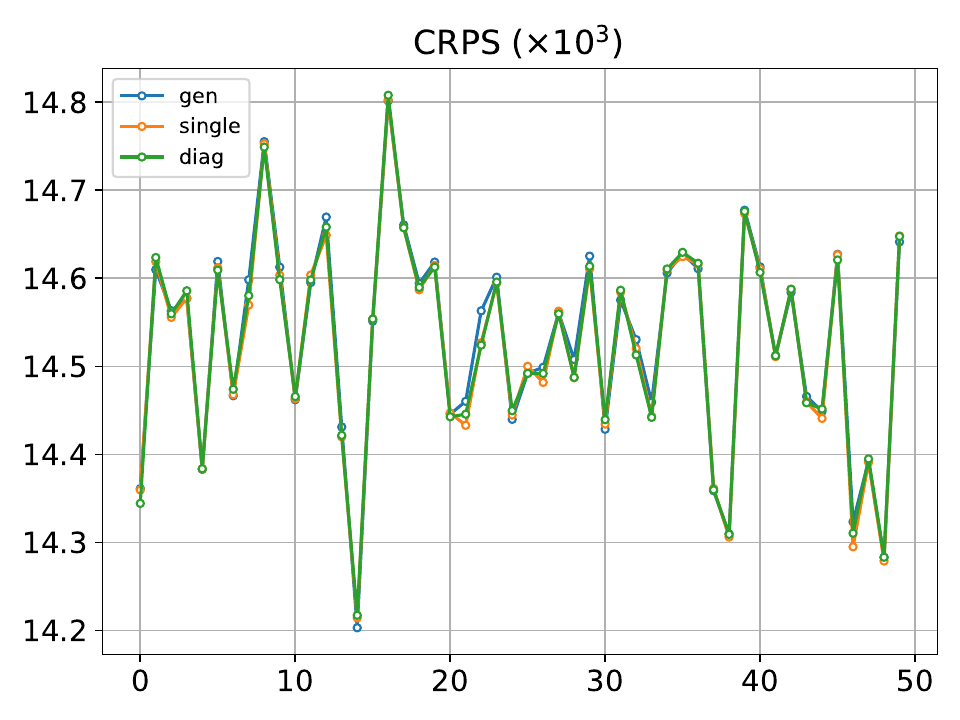}
        \caption{Variable $u$, Std $0.05$}
    \label{fig:crps_200reps_0.05pert_0}
    \end{subfigure}
    \begin{subfigure}{0.3\textwidth}
    \includegraphics[width=\textwidth]{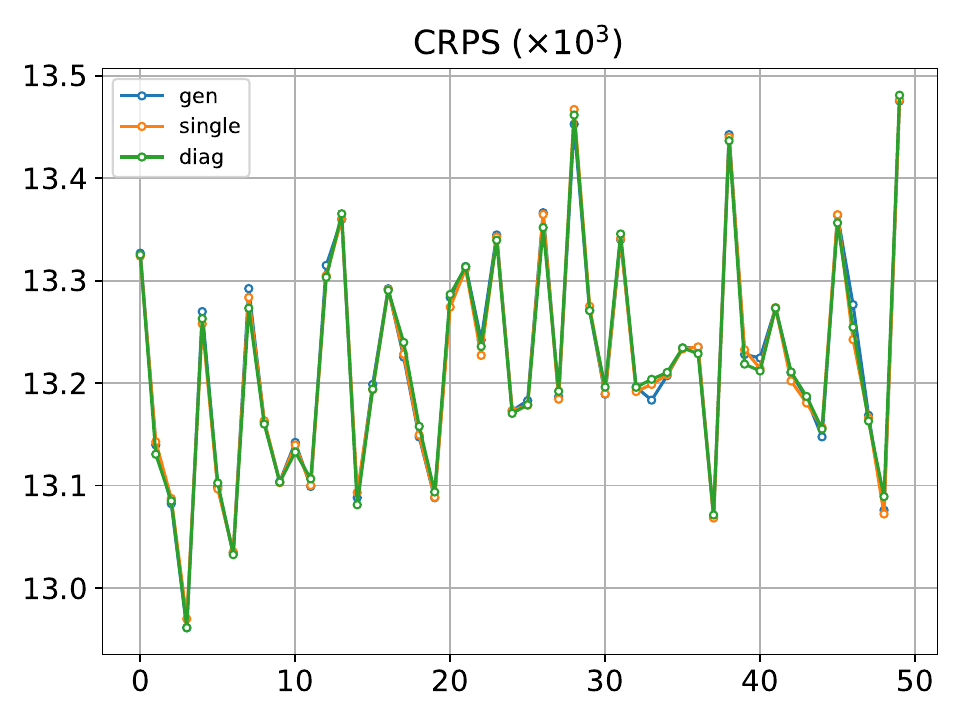}
        \caption{Variable $v$, Std $0.05$}
    \label{fig:crps_200reps_0.05pert_1}
    \end{subfigure}
    \begin{subfigure}{0.3\textwidth}
    \includegraphics[width=\textwidth]{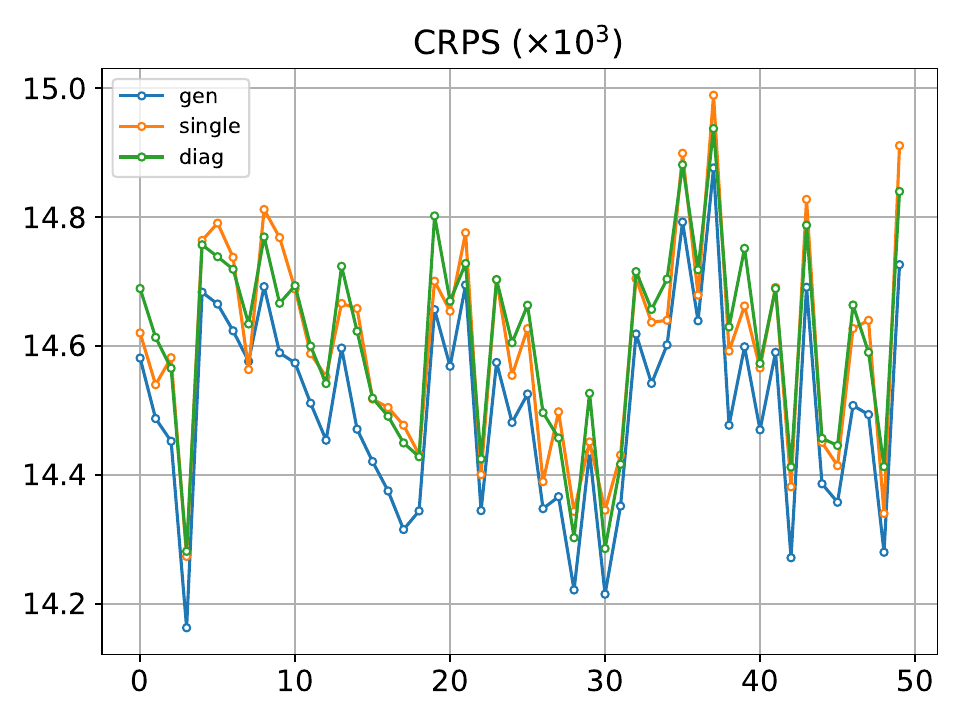}
        \caption{Variable $e$, Std $0.05$}
    \label{fig:crps_200reps_0.05pert_2}
    \end{subfigure}
    \caption{CRPS Scores at forecast times for a lead time of 200 calibration timesteps for all three shallow water variables (u,v, and e). We plot them for different initial noise standard deviations: no noise ($0.0$), small noise ($0.001$) and large noise ($0.05$). The graphs show the results for three different velocity perturbation distributions. Blue lines show the generative model noise, green lines are diagonal gaussian noise with spatially independent variance, and orange is gaussian noise with the same variance in all locations. We observe that small initial noise shows an advantage of the generative model, which fades away as initial uncertainty becomes large. Also, the generative model performs significantly better in the implicit variable e.}
    \label{fig:crps}
\end{figure}

\begin{figure}
    \centering
    \begin{subfigure}{0.3\textwidth}
    \includegraphics[width=\textwidth]{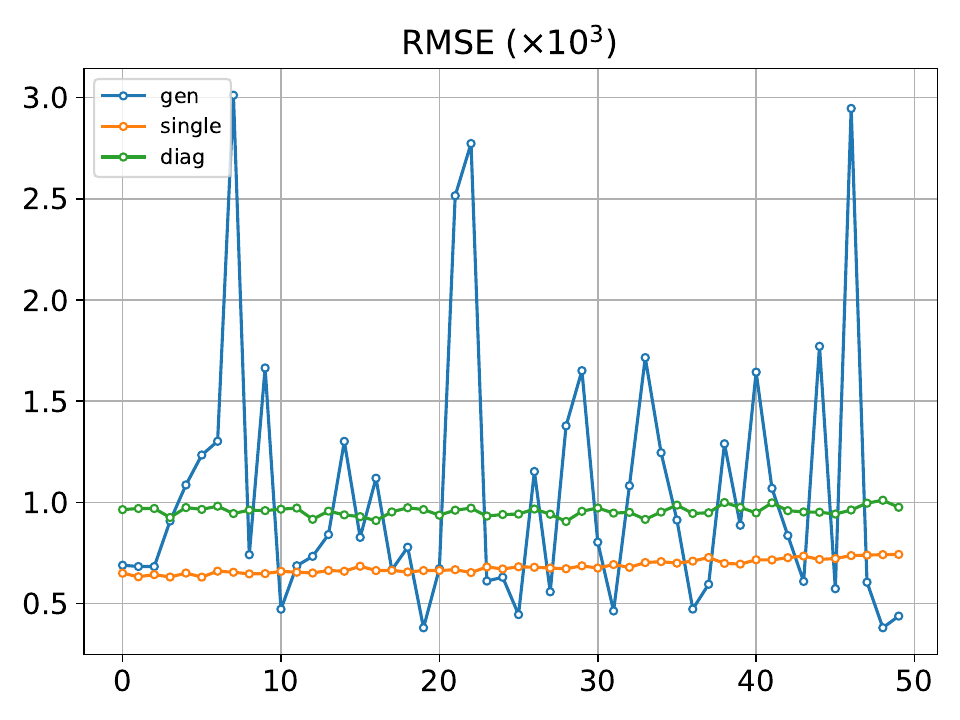}
        \caption{Variable $u$, Std $0.0$}
    \label{fig:rmse_200reps_0}
    \end{subfigure}
    \begin{subfigure}{0.3\textwidth}
    \includegraphics[width=\textwidth]{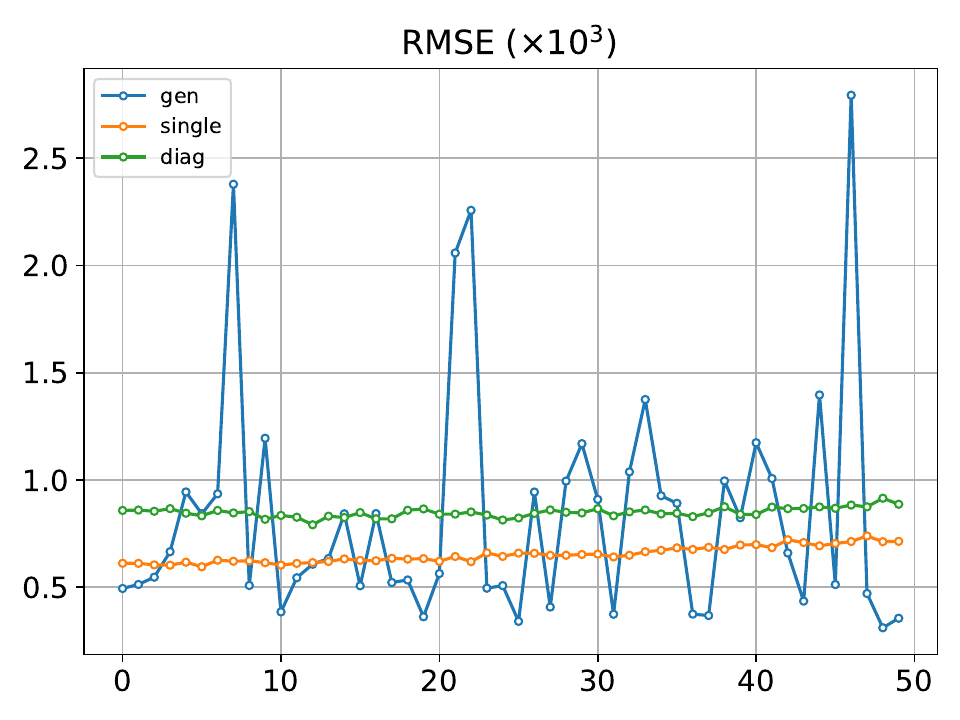}
        \caption{Variable $v$, Std $0.0$}
    \label{fig:rmse_200reps_1}
    \end{subfigure}
    \begin{subfigure}{0.3\textwidth}
    \includegraphics[width=\textwidth]{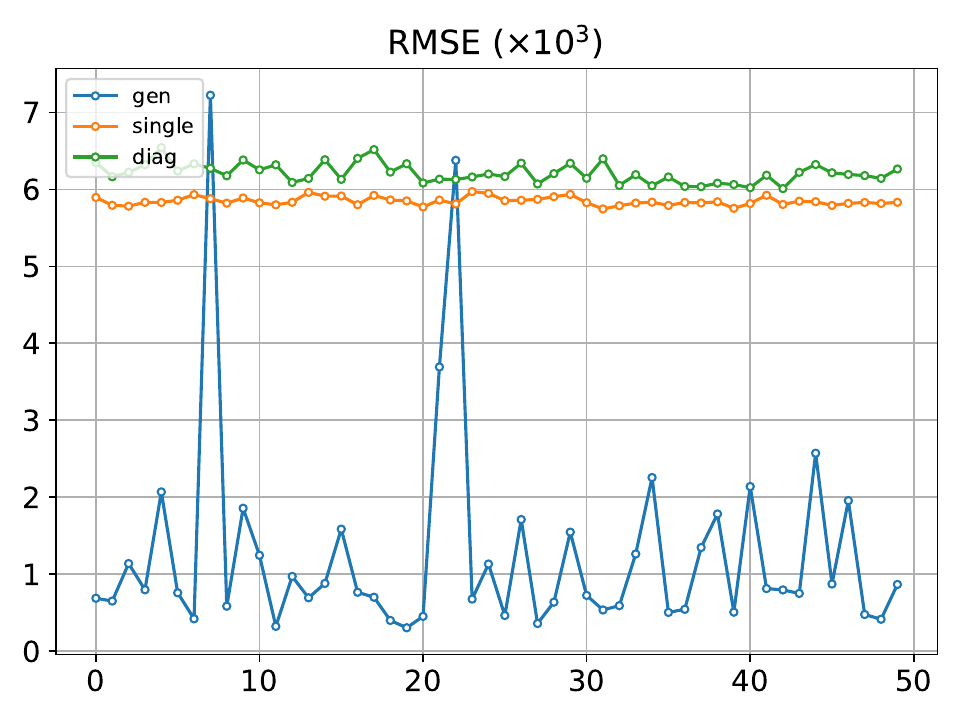}
        \caption{Variable $e$, Std $0.0$}
    \label{fig:rmse_200reps_2}
    \end{subfigure}
    \hfill
    \begin{subfigure}{0.3\textwidth}
    \includegraphics[width=\textwidth]{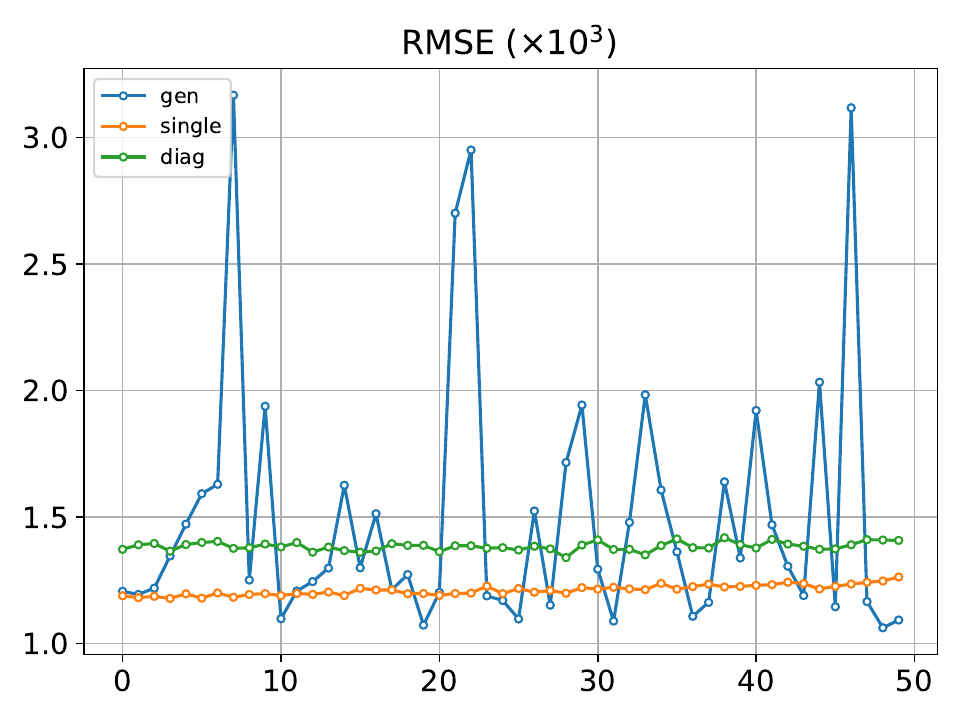}
        \caption{Variable $u$, Std $0.001$}
    \label{fig:rmse_200reps_0.001pert_0}
    \end{subfigure}
    \begin{subfigure}{0.3\textwidth}
    \includegraphics[width=\textwidth]{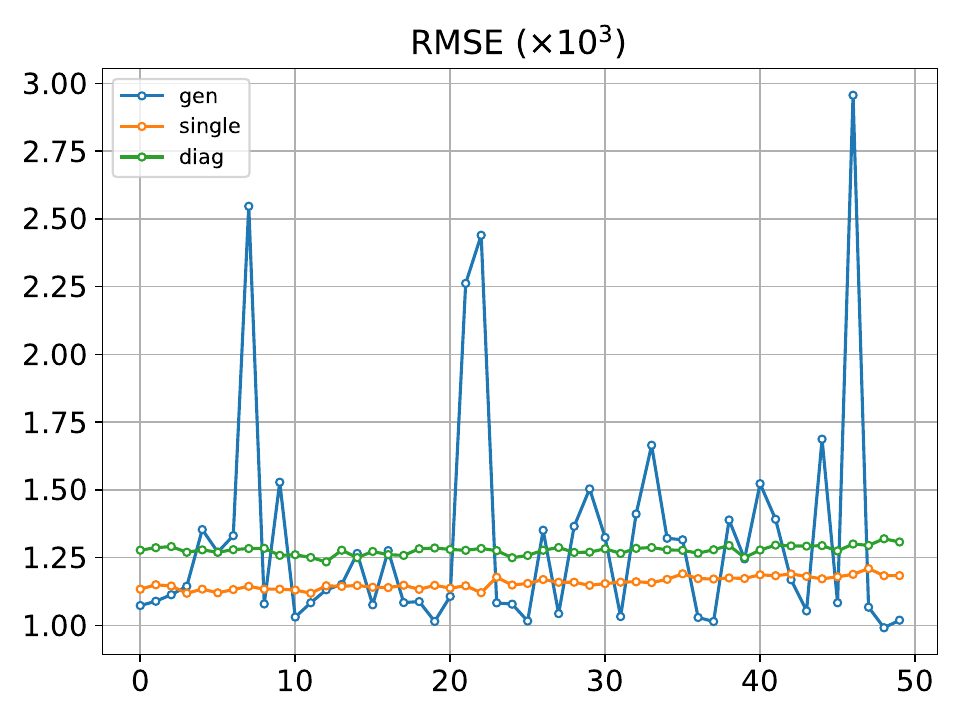}
        \caption{Variable $v$, Std $0.001$}
    \label{fig:rmse_200reps_0.001pert_1}
    \end{subfigure}
    \begin{subfigure}{0.3\textwidth}
    \includegraphics[width=\textwidth]{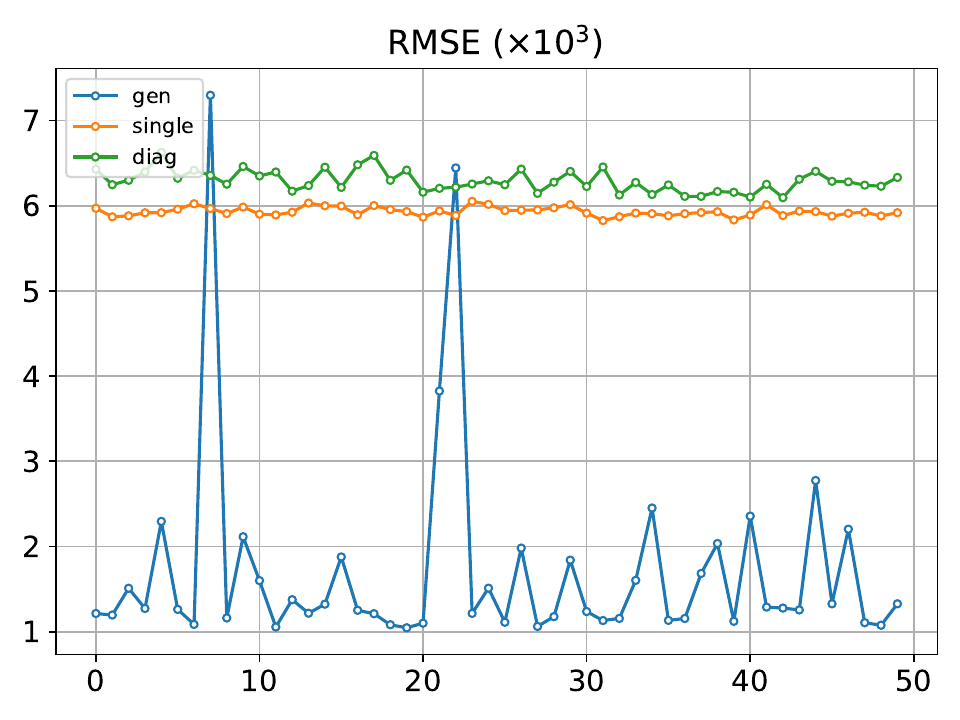}
        \caption{Variable $e$, Std $0.001$}
    \label{fig:rmse_200reps_0.001pert_2}
    \end{subfigure}\hfill
    \begin{subfigure}{0.3\textwidth}
    \includegraphics[width=\textwidth]{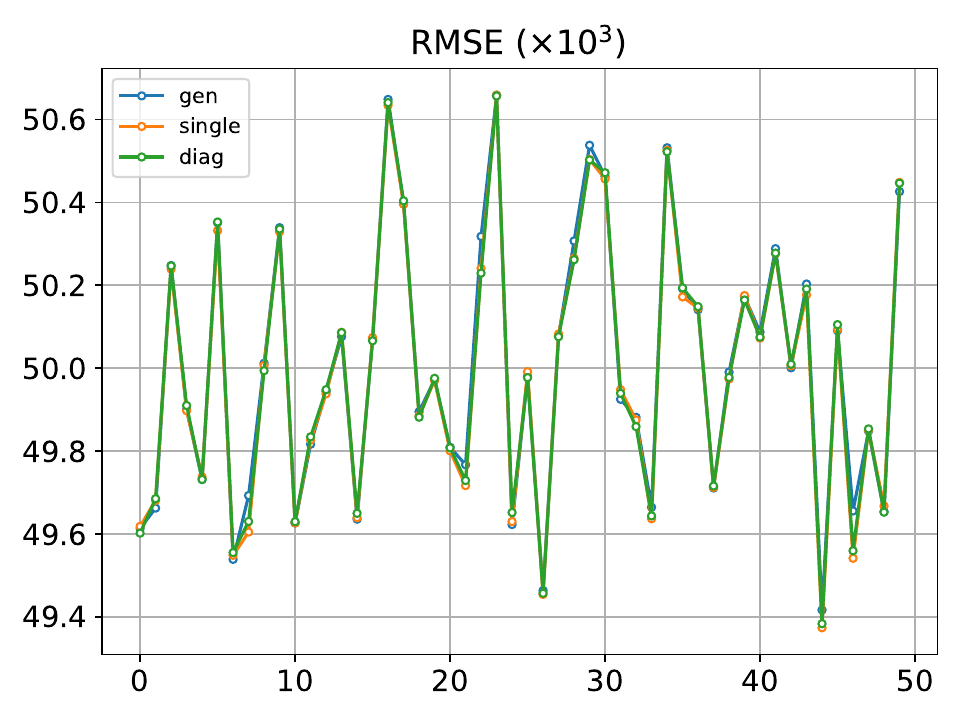}
        \caption{Variable $u$, Std $0.05$}
    \label{fig:rmse_200reps_0.05pert_0}
    \end{subfigure}
    \begin{subfigure}{0.3\textwidth}
    \includegraphics[width=\textwidth]{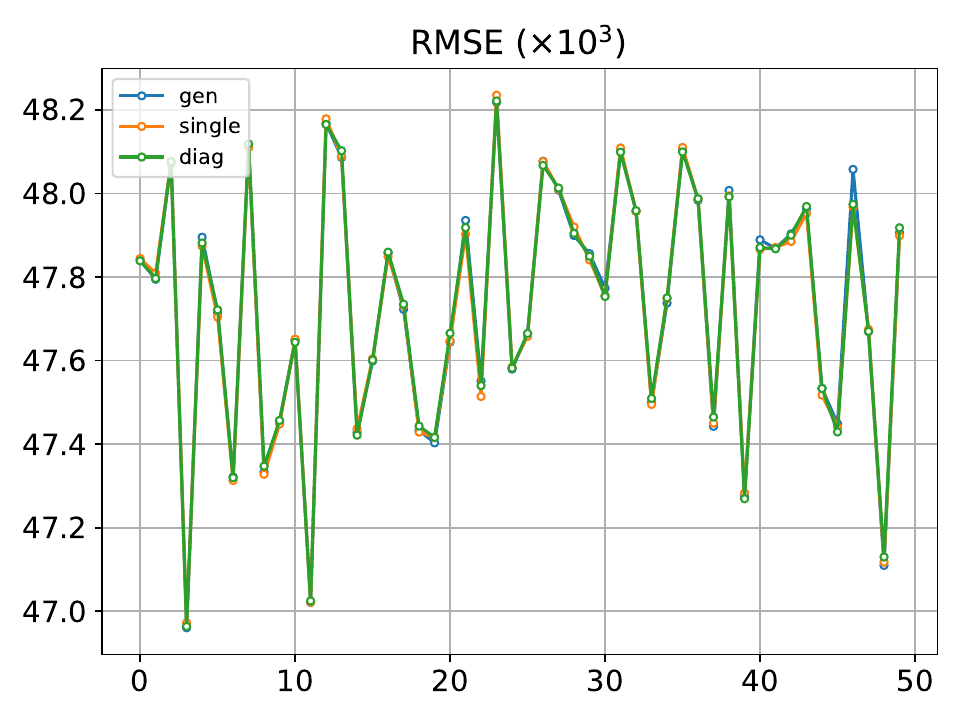}
        \caption{Variable $v$, Std $0.05$}
    \label{fig:rmse_200reps_0.05pert_1}
    \end{subfigure}
    \begin{subfigure}{0.3\textwidth}
    \includegraphics[width=\textwidth]{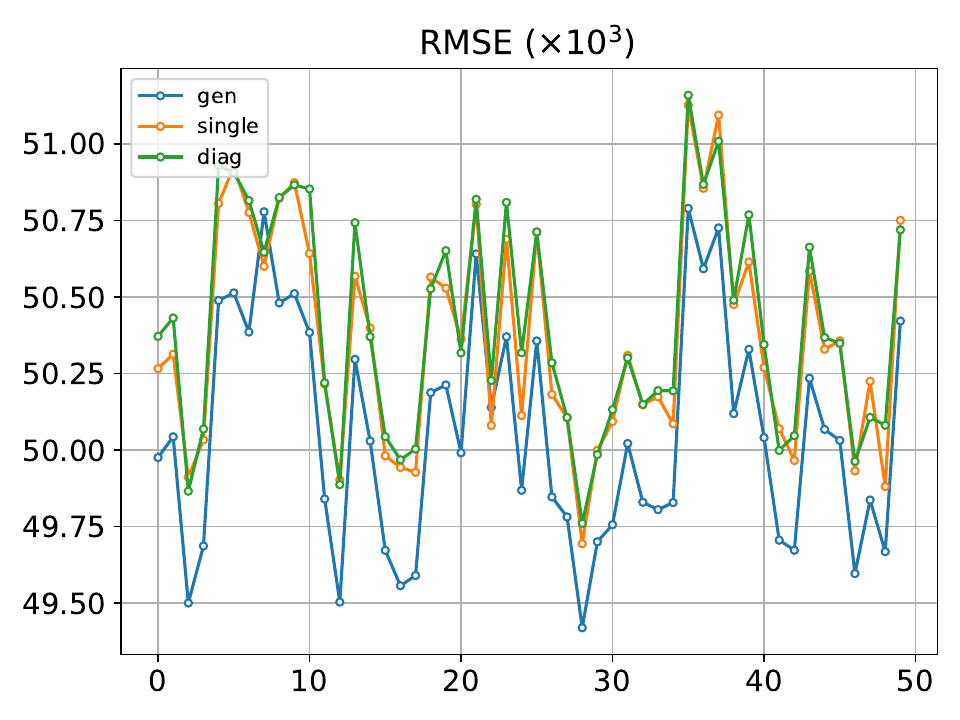}
        \caption{Variable $e$, Std $0.05$}
    \label{fig:rmse_200reps_0.05pert_2}
    \end{subfigure}
    \caption{RMSE scores at forecast times for a lead time of 200 calibration timesteps for all three shallow water variables (u,v, and e). We plot them for different initial noise standard deviations: no noise ($0.0$), small noise ($0.001$) and large noise ($0.05$). The graphs show the results for three different velocity perturbation distributions. Blue lines show the generative model noise, green lines are diagonal gaussian noise with spatially independent variance, and orange is gaussian noise with the same variance in all locations. We observe that small initial noise shows an advantage of the generative model, which fades away as initial uncertainty becomes large. Also, the generative model performs significantly better in the implicit variable e.}
    \label{fig:rmse}
\end{figure}

The results of the CRPS score are depicted in Figure~\ref{fig:crps} below and the results of the RMSE metric are shown in Figure~\ref{fig:rmse}. Both metrics show better forecasts for the generative model in case of no and low initial uncertainty in the ensemble. Especially the forecast results for the height variable are significantly better in the generative noise setting. The results in the u and v variables are somewhat less pronounced.

\begin{figure}
    \centering
    \begin{subfigure}{0.45\textwidth}
    \includegraphics[width=\linewidth]{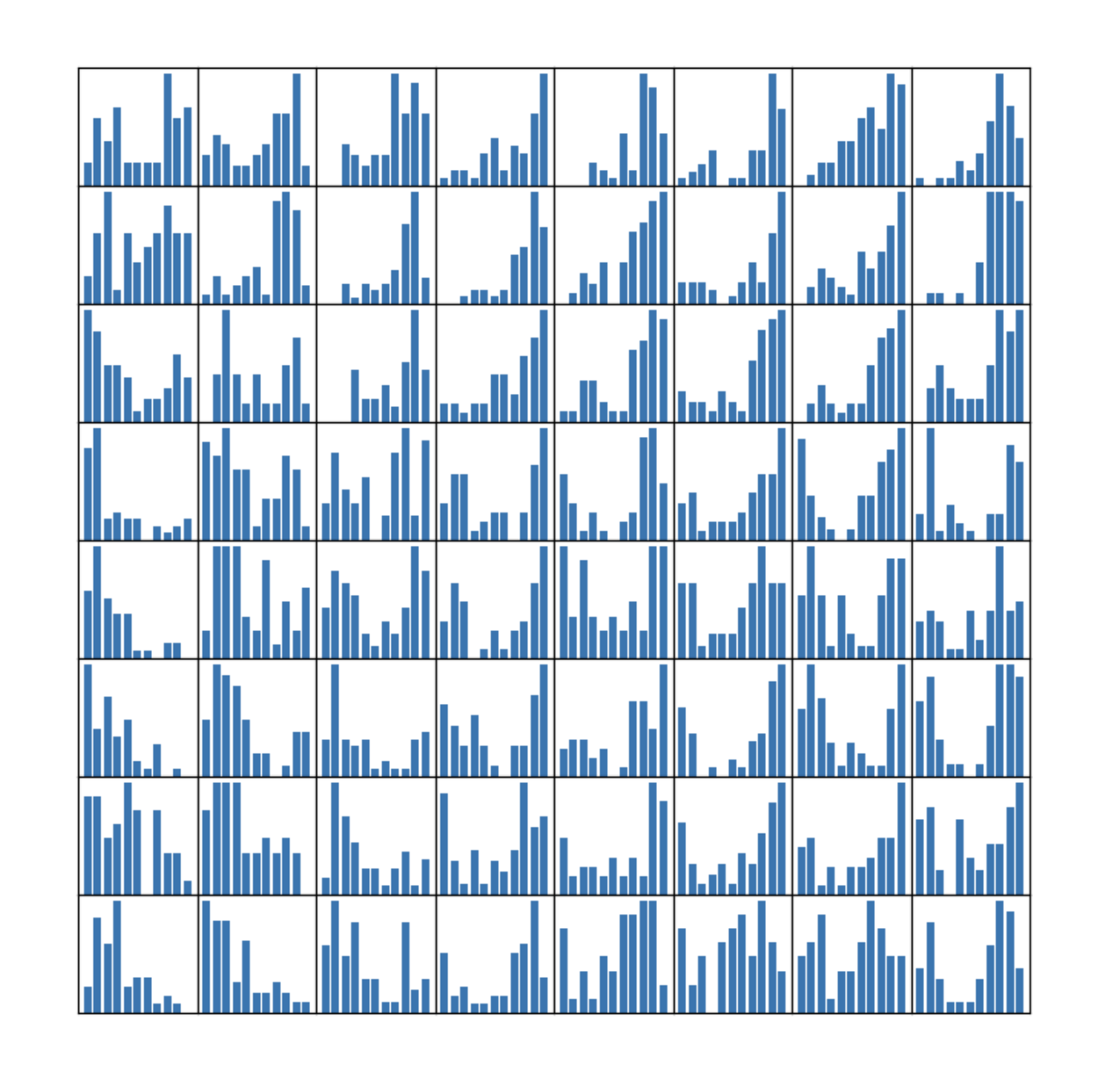}
    \caption{Generative model distribution}
    \label{fig:rh-gen}
    \end{subfigure}
    \begin{subfigure}{0.45\textwidth}
    \includegraphics[width=\linewidth]{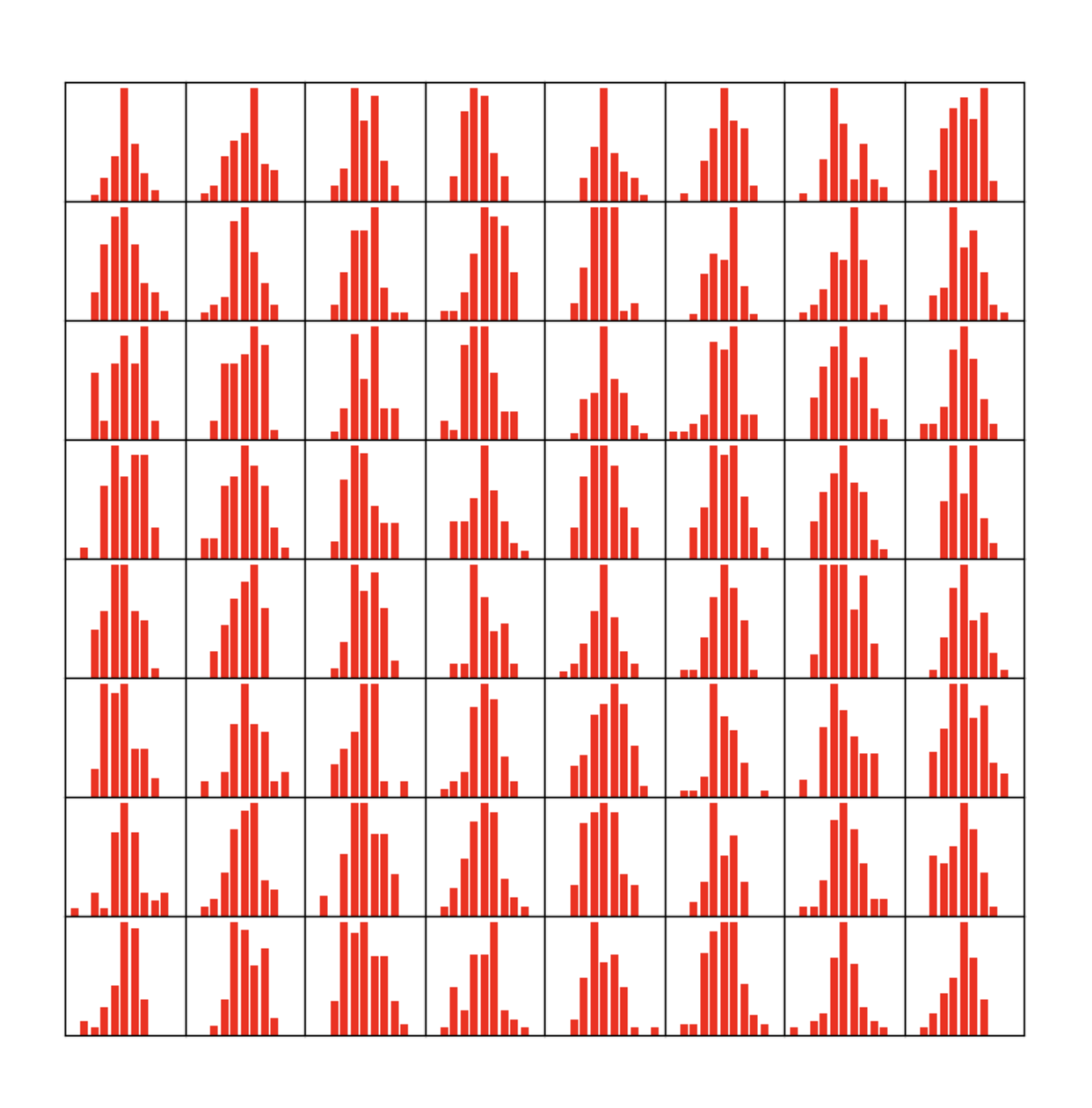}
    \caption{Gaussian distribution}
    \label{fig:rh-gauss}
    \end{subfigure}
    \caption{Rank histograms. (a) Rank Histograms of the generative noise ensemble with a given forecast horizon. (b) Rank Histograms of the Gaussian noise ensemble with a given forecast horizon.}
    \label{fig:rank-histograms}
\end{figure}

The forward run ensembles are also assessed for a longer duration with ensembles of 10 particles for $1,000$ calibration steps, without resetting to the truth. We produce rank histograms from those runs using repetitions with different noise samples. Here, we compare the generative model noise to a Gaussian with the same overall mean and variance as the generative model. The results are shown in Figure~\ref{fig:rank-histograms}. Figure~\ref{fig:rh-gen} shows the rank histograms for the Gaussian noise which are overall more uniform than the clearly overdispersed Gaussian ensembles in~\ref{fig:rh-gauss}.

\begin{figure}
    \centering
    \includegraphics[width=0.7\linewidth]{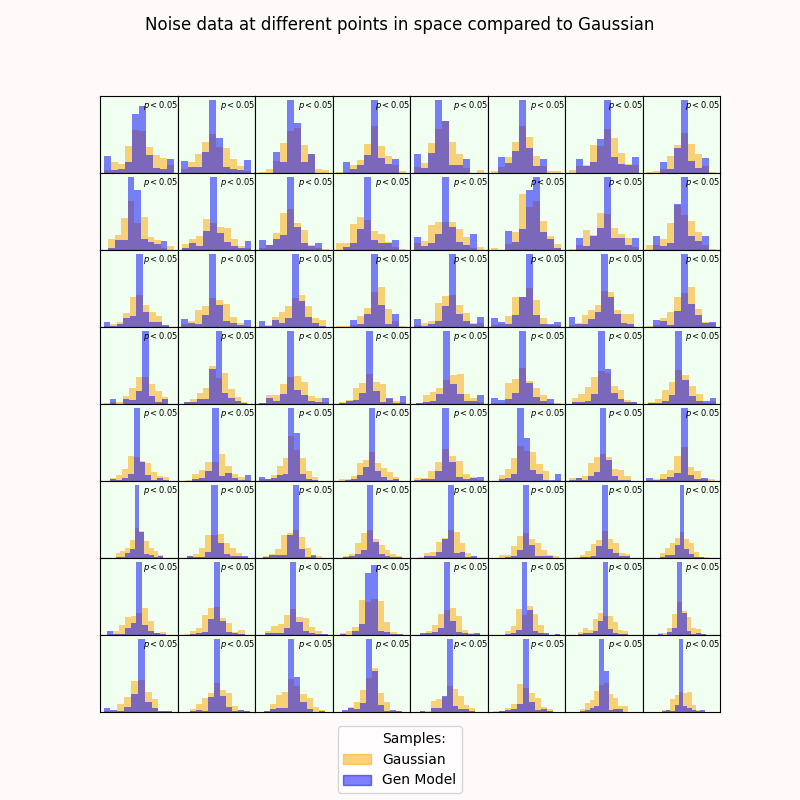}
    \label{fig:data_hist}
    \caption{ Distribution of the generated noise values compared to Gaussians at every pixel in a central region. A Kolmogorov-Smirnov one-sample test has been performed to check if the data come from a normal distribution. The hypothesis was rejected for all locations indicating that the generated noise does not come from a simple normal distribution.}
\end{figure}

\begin{figure}
    \centering
    \includegraphics[width=0.6\linewidth]{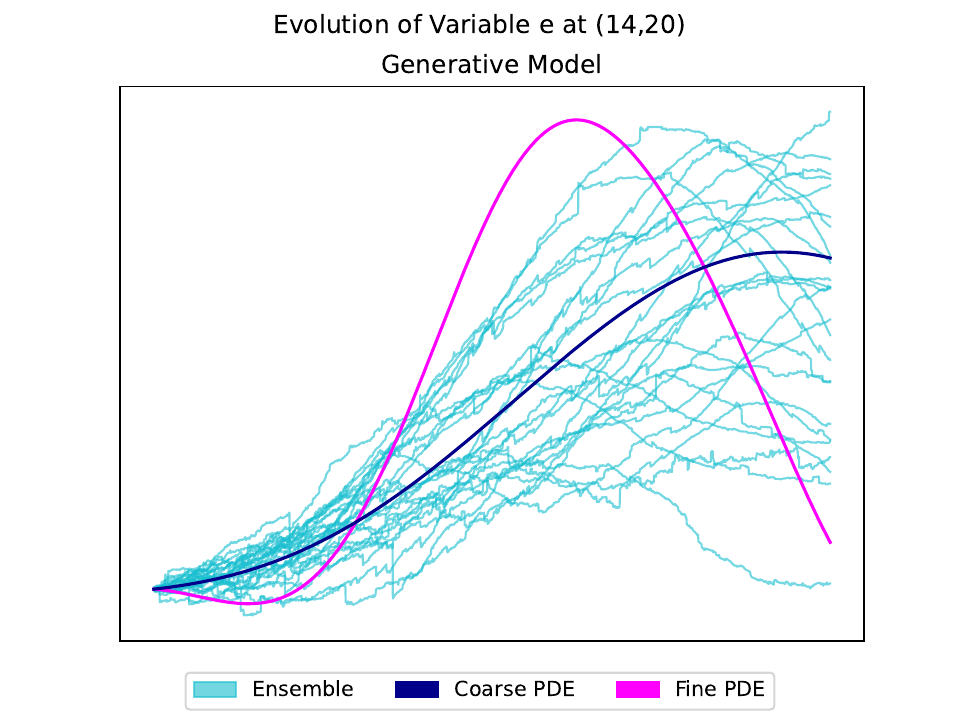}
    \caption{Ensemble plot for forward runs of the SPDE with generative model noise. We compare the evolution of the fine scale PDE projected onto the coarse grid with the evolution of the PDE run on the coarse grid and an ensemble with generative noise at a central grid location. Horizontal axis is time.}
    \label{fig:forward-runs}
\end{figure}

\section{Conclusions and Future Work}
\label{sec:conclusion}

In this work the feasibility of using modern generative models for the generation of appropriate noise distributions in stochastic models for subgridscale effects in fluid dynamics has been investigated.
To this end we implemented a Diffusion Schr\"odinger Bridge model for the generation of Rotationg Shallow Water noise and performed a comparative study of the generated ensemble in terms of established forecast metrics, RMSE and CRPS.
The results show that the generative model samples display an advantage over the gaussian noise ensemble in the case of low initial uncertainty. This result indicates that the generative model is more effective at capturing the fine scale effects on the coarse dynamics than an Gaussian noise ensemble. 

In future work, we will compare the generative model noise against a model using a Karhunen-Loeve decomposition of the dataset, according to the previously developed method in~\cite{us}. Moreover, studies on different underlying fluid models need to be performed in addition to this initial study on the rotating shallow water equations.  

\vspace{5mm}
\noindent\textbf{Funding}\\
\noindent All three authors have been supported by the European Research Council (ERC) under the European Union’s Horizon 2020 Research and Innovation Programme (ERC, Grant Agreement No 856408). 

\vspace{3mm}
\noindent\textbf{Data availability statement} \\
\noindent Data sharing not applicable to this article as no datasets were generated or analysed during the current study.

\vspace{3mm}
\noindent\textbf{Conflict of interest statement} \\
On behalf of all authors, the corresponding author states that there is no conflict of interest.

\bibliographystyle{plain}
\bibliography{references}
\end{document}